\documentclass{article}

\usepackage{cite}

\usepackage{graphicx} 
\usepackage{amsmath} 
\usepackage{amssymb}  
\usepackage{float}
\usepackage{multicol}
\usepackage{subfigure}
\usepackage{tabularx,colortbl}
\usepackage{multirow}
\usepackage{tikz}

\definecolor{trueColor}{RGB}{0,102,0}
\definecolor{falseColor}{RGB}{204,0,0}

\title{\LARGE \bf
 Learning Inverse Statics Models Efficiently
}

\newcommand{\ve}[1]{\mathbf{#1}}

\newcommand{\diag}[0]{\mbox{diag}\xspace}

\newcommand{\ma}[1]{\mbox{\boldmath$#1$}}

\usepackage{blindtext}
\usepackage{todonotes}

\usepackage{algorithm}
\usepackage[end]{algpseudocode} 
\usepackage{cite}

\algnewcommand\algorithmicinput{\textbf{INPUT:}}
\algnewcommand\INPUT{\item[\algorithmicinput]}

\algnewcommand\algorithmicoutput{\textbf{OUTPUT:}}
\algnewcommand\OUTPUT{\item[\algorithmicoutput]}

\author{ Rania Rayyes, Daniel Kubus, Carsten Hartmann and Jochen Steil
	\thanks{The authors are with Technische Universit{\"a}t Braunschweig, Institut f{\"u}r Robotik und Prozessinformatik, 38106 Braunschweig, Germany
		{\tt\small \{rrayyes,dku,jsteil\}@rob.cs.tu-bs.de}}%
}

\begin{document}
\addtolength{\parskip}{-0.25mm}
\setlength{\textfloatsep}{0.1cm}
\maketitle
\thispagestyle{empty}
\pagestyle{empty}


\begin{abstract}
Online Goal Babbling and Direction Sampling are recently proposed methods for direct learning of inverse kinematics mappings from scratch even in high-dimensional sensorimotor spaces following the paradigm of "learning while behaving". 

To learn inverse \emph{statics} mappings -- primarily for gravity compensation -- from scratch and without using any closed-loop controller, we modify and enhance the Online Goal Babbling and Direction Sampling schemes. Moreover, we exploit symmetries in the inverse statics mappings to drastically reduce the number of samples required for learning inverse statics models.


Results for a 2R planar robot, a 3R simplified human arm, and a 4R humanoid robot arm clearly demonstrate that their inverse statics mappings can be learned successfully with our modified online Goal Babbling scheme. Furthermore, we show that the number of samples required for the 2R and 3R arms can be reduced by a factor of at least $8$ and $16$ resp. -- depending on the number of discovered symmetries.


\end{abstract}

\section{Introduction}

Compensating forces and torques due to gravity is an essential ingredient for advanced (model-based) robot control approaches. Usually, gravitational terms of the inverse dynamics models are computed from the Computer-Aided-Design~(CAD) data of the robot or based on estimated link masses and centers of mass~(COMs). If robots with elastic links and/or elastic joints are considered or no prior information on the link masses and COMs is available, \emph{learning} these gravitational terms is the most promising option.
Therefore, we propose and evaluate approaches which learn inverse statics models \emph{online, from scratch} and do not require any prior knowledge or feedback controller.

Learning from scratch is at the core of the popular Goal Babbling scheme which is inspired by infants learning motor skills and has been shown to approximate inverse \emph{kinematics} mappings~(IKMs) within few hundred samples without any prior knowledge~\cite{GB1}. 
To apply Goal Babbling, a set of predefined targets, e.g. a set of positions to be reached scattered on a grid, is required which is then used to obtain a locally valid IKM.
Direction Sampling~\cite{Rolf2013_DS} has recently been proposed to overcome the need for predefined targets and gradually discovers the entire workspace. In contrast to Goal Babbling, targets are generated while exploring and the IKM is learned simultaneously.

While these previous approaches focused on IKMs only, in this paper, we focus on learning inverse \emph{statics} mappings~(ISMs) by modifying the previously proposed online Goal Babbling and Direction Sampling schemes. Several crucial modifications and extensions are required as ISMs have fundamentally different characteristics compared to IKMs:

Firstly, IKMs often map from a lower dimensional domain (observation space) to a higher dimensional codomain (action space) whereas the dimensionality of the domain and codomain in ISMs are usually identical -- since ISMs map from joint space to motor space. 

\begin{figure}[htpb]
	\centering
	\subfigure[Typical inv. kin. mapping]{\label{fig:IKM} \includegraphics[ width=0.4\linewidth]{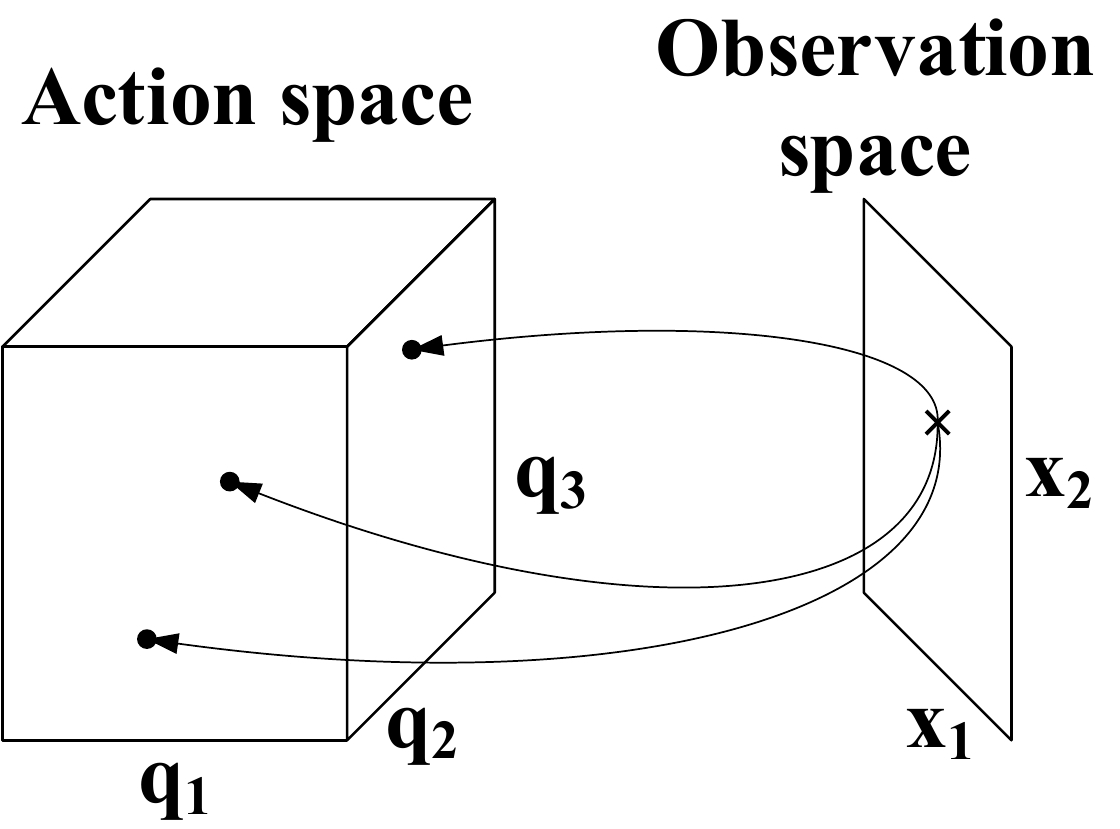}}
	\subfigure[Typical inv. statics mapping]{\label{fig:ISM} \includegraphics[ width=0.48\linewidth]{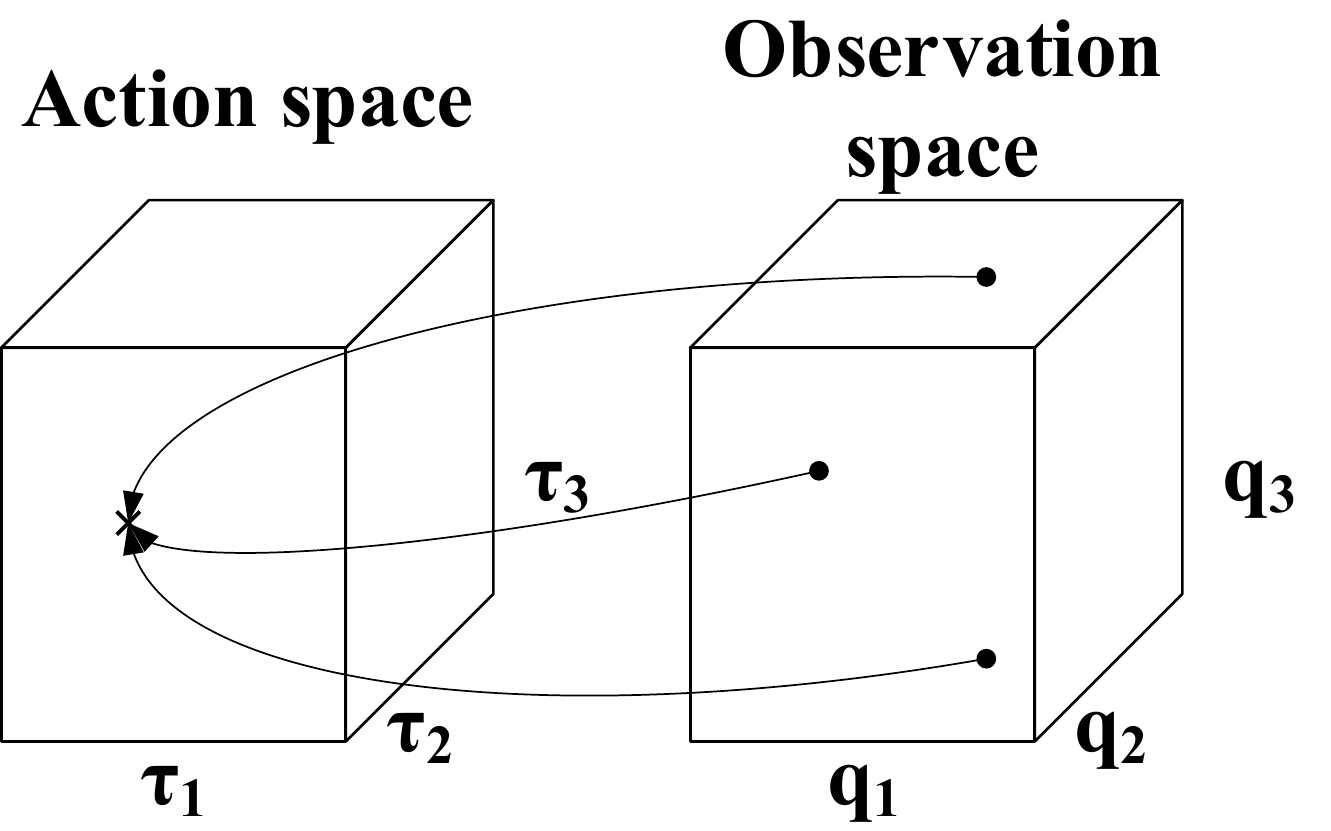}}
	\caption{Characteristics of IKMs and ISMs. In contrast to IKMs which are one-to-many mappings, ISMs are many-to-one mappings. }
	\label{fig:IKMISM}
\end{figure}

As illustrated in Fig.\ref{fig:IKMISM}, IKMs are usually one-to-many mappings, i.e.  each pose can be reached by multiple configurations whereas ISMs are many-to-one mappings, i.e. multiple configurations require the same generalized force to be maintained. 
We propose an approach to exploit this many-to-one characteristic, denoted as \emph{symmetries} of ISMs,  to drastically accelerate learning. More precisely, after discovering functional relations between configurations requiring similar generalized forces, we use these relations to associate the applied generalized force with multiple configurations instead of only the current one -- thus reducing the number of required training samples.
This general property of ISMs can be exploited by any learning scheme and hence is neither specific to Goal Babbling/Direction Sampling nor to online/offline learning approaches in general.

Secondly, in contrast to learning IKMs with Goal Babbling, in ISMs, exploratory noise in action space can result in inadmissible generalized forces, which can -- for instance -- cause the robot to hit its joint limits and hence severely disturb the learner. Thus, a set of admissible generalized forces has to be estimated or learned. 

Thirdly, the employed weighting scheme has to be modified in order to learn ISMs with online Goal Babbling.

The remainder of the paper is structured as follows:
The following section reviews related work. Sec.~\ref{sec:GB} addresses the proposed modifications of online Goal Babbling and Direction Sampling for learning ISMs.
Sec.~\ref{sec:Sym} introduces the notions of primary and secondary symmetries in ISMs and discusses their discovery and exploitation in learning. 
Sec.~\ref{sec:expermints} presents experimental results for modified online Goal Babbling for 2R, 3R, and 4R manipulators. Moreover, the efficiency gained by exploiting symmetries for learning ISMs is illustrated by Direction Sampling (online) and a batch learning technique for a 2R and a 3R manipulator resp.

\section{Related Work}
Movement skill learning and motion control are attractive topics in neuroscience and machine learning in order to describe and replicate human development and learning processes. A variety of learning schemes has been proposed to imitate and replicate human motor control based on internal models~\cite{Miall}, i.e. forward models (which map motor commands to predicted outcomes) and inverse models (which map desired outcomes to suitable motor commands).

Popular approaches for learning inverse models are distal teachers~\cite{Jordan92forwardmodels:} and feedback error learning~\cite{Kawato}. 
However, these approaches require a forward model or a feedback controller resp. in order to learn the inverse mapping. 

Learning forward models by Motor Babbling was proposed in~\cite{Demiris:2005}. Training samples are acquired by \emph{random} exploration of motor commands and the action-outcome mapping is learned. 

However, recent studies proved that even infants do not behave randomly as implied by Motor Babbling; instead, they demonstrate goal-directed motion~\cite{von} already few days after birth. 
For example, they learn how to reach objects by trying to reach them, and they adapt their motion by iterating their trials. Such intrinsically motivated goal-directed learning approaches to obtain IKMs were proposed in~\cite{Rolf2010} and the term Goal Babbling was coined.
Apart from Goal Babbling various other schemes for direct~\cite{Souza,Thuruthel:162} and incremental~\cite{Schaal,Oudeyer2013} learning of IKMs were proposed. 

In~\cite{GB1}, an incremental \emph{online} learning scheme based on the original Goal Babbling concept was proposed which scales well to high dimensions in a ”learning-while-behaving” fashion, e.g. up to a 50R planar arm~\cite{GB1}, up to a 9 DoF humanoid robot~\cite{GB}, and is also applicable to soft robots~\cite{Ro1f14}.
Direction Sampling was proposed as an extension of online Goal Babbling~\cite{Rolf2013_DS} in order to explore the entire workspace and learn the IKM simultaneously. The applicability and performance of Direction Sampling have been demonstrated in 2D for up to a 50R planar arm~\cite{Rolf2013_DS} and in 3D for a 9 DoF humanoid robot~\cite{GB}.

In contrast to the latter approaches, which only considered IKMs, we investigate if goal-directed learning schemes based on online Goal Babbling and Direction Sampling can be devised to successfully and efficiently learn ISMs from scratch online. To this end, we propose and evaluate extensions and several modifications of online Goal Babbling and Direction Sampling which enable their application for learning ISMs -- primarily for gravity compensation.

Early data-driven gravity compensation schemes used iterative procedures to facilitate end-point regulation~\cite{DeLuca:94,DeLuca:96}.
Recent works~\cite{Giorelli:15,Thuruthel:16} have explored data-driven learning techniques to tackle the inverse statics problem for end-point control of continuum manipulators in task space. However, simple motor babbling and feedback controllers were applied to obtain the training data and learning was performed fully offline.
In contrast to these approaches, our online goal-directed learning schemes gradually explore the configuration space and learn the ISM from scratch while exploring. 
Moreover, we exploit symmetries of ISMs to drastically reduce the number of required training samples.


%



\section{Modifying Online Goal Babbling and Direction Sampling for Learning ISMs}\label{sec:GB}
As our primary target are manipulators with revolute joints only, we will use the term torques instead of generalized forces in the remainder of this paper.

Let $\mathcal{Q}_v$ denote the open set of permissible configurations $\ve{q}$, and $\mathcal{T}_s$ the corresponding static torques for these configurations.
In this paper, we aim to learn the mapping
$\ve{G}:\mathcal{Q}_v\rightarrow \mathcal{T}_s$
that assigns a torque $\ve{\tau} \in \mathcal{T}_s$ to each valid configuration $\ve{q}\in\mathcal{Q}_v$ which is required to maintain this configuration $\ve{q}$.
To this end, we employ the powerful online Goal Babbling~\cite{GB1} and Direction Sampling~\cite{Rolf2013_DS} schemes. However, to apply them successfully to bootstrap ISMs, several modifications to the original schemes are necessary:

\subsection{Modified Online Goal Babbling Scheme}
In our modified online Goal babbling scheme, we identify the action space $\mathcal{A}$ with the vector space of torques and the observation space $\mathcal{O}$ with the configuration space of the robot. 
At the beginning, the initial inverse estimate $\ve{\hat{G}}$ at $t = 0$ suggests some default torque $\ve{\hat{G}}(\ve{q^*}) = const = \ve{\tau}^{home}$ corresponding to some comfortable default configuration (home posture) $\ve{q}^{home}$ for any configuration $\ve{q}^*$. The goal is to learn how to maintain some predefined set of goals $\mathcal{Q}^*$ in configuration space. The goals in $\mathcal{Q}^*$ are chosen randomly and iteratively, and continuous paths of intermediate targets $\ve{q}_t^*$ are generated between the goals in $\mathcal{Q}^*$. $\ve{q}^{home}$ will be selected as a target with a probability of $p^{home}\approx0.01\%$. 

The agent tries to reach each $\ve{q}_t^*$ using the online Goal Babbling basic scheme (GBSCHEME, cf.~Alg.~\ref{algo_gb}) as following:
The current inverse estimate is used as a motor torque
$\ve{\hat{\tau}}^{*}_{t}$, and correlated exploratory noise $\ve{\sigma}$~\cite{GB1} is added in order to discover and learn new configurations as specified in Eq.~(\ref{eq:gb1}):
\begin{equation}\label{eq:gb1}
\ve{\tau}^+_t=\ve{\hat{\tau}}^*_t+\ve{\sigma}(\ve{q}^*_t,t)
\end{equation}
where $t$ is the time step, and $\ve{\tau}^+_t$ is the torque applied to the robot. The outcomes $(\ve{\tau}^{+}_{t},\ve{q}^{+}_{t})$ are observed and the inverse estimate is updated based on this sample immediately before the next intermediate target is generated. This allows to unfold the inverse estimate from the home configuration and finally provide suitable torques for all $\mathcal{Q}^*$. 
As an incremental regression mechanism is required, a Local Linear Map (LLM)\cite{GB1} is used in our modified online Goal Babbling scheme.

\subsubsection{Set of Static Torques}
\label{sec:SOST}
The exploration has to be limited to the set of torques that correspond to valid configurations of the robot. This set of static torques $\boldsymbol{SST}$
\begin{equation}
\mathcal{T}_s=\left\{\ve{\tau}|\exists \ve{q}\in \mathcal{Q}_v:\ve{\tau}-\ve{G}(\ve{q})=0\right\}
\end{equation}
denotes the set of torques required to maintain configurations $\ve{q}\in \mathcal{Q}_v$.
Due to exploratory noise $\ve{\sigma}$, some torques $\ve{\tau}_t^+ \notin \mathcal{T}_s$ would be applied to the robot and thus the resulting configuration $\ve{q}$ will not be in $\mathcal{Q}_v$. Since $\ve{\tau}_t^+ \notin \mathcal{T}_s$ would disturb the learner due to invalid configurations, e.g. the robot hits its joint limits, such torques should be avoided during bootstrapping. 

An obvious strategy is to explore the $\boldsymbol{SST}$ beforehand and then exploit this knowledge in the bootstrapping process. To save time, this exploration can be performed in conjunction with symmetry discovery as detailed in Sec.~\ref{sec:acc_learn_sym}.
After $\boldsymbol{SST}$ exploration, Delaunay triangulation is used to estimate its boundary.

After adding exploratory noise $\ve{\sigma}$ to the learner output $\ve{\hat{\tau}}^*_t$ in order to explore the action space, the resulting torque $\ve{\tau}_t^+$ will be applied to the robot if $\ve{\tau}_t^+\in\mathcal{T}_s$ holds or (if $\ve{\tau}_t^+\notin\mathcal{T}_s$) it will be assigned to the nearest vertex on the boundary of the $\boldsymbol{SST}$ (cf.~l.~16 in Alg.\ref{algo_gb}) as illustrated in Sec.~\ref{sec:expermints}.

\subsubsection{Sample Weighting and Selection Scheme} 
In order to favor samples used in the training phase, one criterion of the weighting scheme, which has been previously proposed in~\cite{GB1}, is adopted:
\begin{equation}
w_t^{dir}=\dfrac{1}{2}(1+cos\sphericalangle (\ve{q}_t^*-\ve{q}_{t-1}^*,\ve{q}^+_t-\ve{q}^+_{t-1})
\end{equation}
The $w_t^{dir}$ direction criterion assesses whether the observed configuration and the intended one align well. 
The weights of the samples are then updated to minimize the weighted squared error $E_t$ given in Eq.~(\ref{eq:werror})~\cite{GB1}:
 \begin{equation}\label{eq:werror}
 E_t = w_t^{dir} \lVert{\ve{\tau}^+_t-\ve{\hat{\tau}}_t^+}\lVert^2
 \end{equation}
Note that executing $\ve{\tau}_t^+$ will result in $\ve{q}_t^+$ and the corresponding torque estimated by the learner for $\ve{q}_t^+$ is denoted by $\ve{\hat{\tau}}_t^+$. Hence, the goal is to minimize the error $E_t$ to improve the estimation accuracy.


Alg.~\ref{algo_gb} summarizes the modified online Goal Babbling algorithm. For a selected number of iterations and a set of targets $\mathcal{Q}^*$, intermediate targets are generated and the goal babbling basic scheme (GBS\textsc{CHEME}) is applied. 

\begin{algorithm}[!tb]
	\caption{Modified Online Goal Babbling}\label{algo_gb}
	\begin{algorithmic}[1]
	
		\Procedure{GoalBabbling} {$\ve{\tau}^{home}$,$\mathcal{Q}^*$,$\boldsymbol{SST}$}
		\State initialize learner $\ve{\hat{G}}(\ve{q}^{*}) = \ve{\tau}^{home}$ 
		\State \textbf{for $N$ number of iterations} 
		\State \indent \textbf{for} each goal $\ve{\varrho^*}\in\mathcal{Q^*}$
		\State \indent \indent \textbf{for} $l=1:L$
		\State \indent \indent \indent generate an intermediate target $\ve{q}_t^*$
		\State \indent \indent \indent \Call{GBScheme}{$\ve{q}_t^*$}
		\State \indent \indent \textbf{end for}
		\State  \indent \textbf{end for}
		\State  \textbf{end for}
		\EndProcedure
		\Procedure{GBScheme}{$\ve{q}_t^*$}
		\State  estimate torque value $\ve{\hat{\tau}}_{t}^{\ast}$ for {$\ve{q}_t^*$}
        \State add exploratory noise $\ve{\sigma}$:\newline \indent \indent  $\ve{\tau}^+_t=\ve{\hat{\tau}}_{t}^{\ast} +\ve{\sigma}(\ve{q}_t^*,t)$ 
		\State \textbf{if} $\ve{\tau}_t^+ \notin \boldsymbol{SST}$
		\State \indent   set $\ve{\tau}^+_t=\ve{\tau}_m$ where \{$\ve{\tau}_m\in\mathcal{T}_s: \forall \ve{\tau}_n \in \mathcal{T}_s \newline \indent \indent dist(\ve{\tau}^+_t,\ve{\tau}_m) \leqslant dist(\ve{\tau}^+_t,\ve{\tau}_n)$\}
		\State  \textbf{end if}
		\State execute $\ve{\tau}^+_t$ and observe $\ve{q}^+_t$
		\State compute weight $w_t^{dir}$
	    \State $learner \longleftarrow (\ve{\tau}^+_t,\ve{q}^+_t,w_t^{dir})$
	    \EndProcedure
	\end{algorithmic}
\end{algorithm}

\subsection{Modified Direction Sampling for Learning ISMs}\label{sec:DS}
Modified Direction Sampling employs modified online Goal Babbling to learn ISMs and to discover the permissible configurations simultaneously without predefined goals. Instead the goals will be generated during the learning process. 

Alg.~\ref{algo_ds} shows the individual steps of our modified Direction Sampling approach.
The robot starts from its home posture $\ve{q}^{home}$ and the targets are generated along a random direction $\ve{\varDelta q}$  as given in Eq.~(\ref{e1}):

	\begin{equation}\label{e1}
	{\ve{q}^*_t} = {\ve{q}^*_{t-1}} + {\dfrac{\varepsilon}{\lVert{\ve{w}^T\ve{\varDelta q}}\rVert}}\cdot \ve{\varDelta q}
	\end{equation}
	
	where $\ve{q}_t^*$ is the current generated target, $\ve{q}_{t-1}^*$ is the previous one, $\ve{w}$ is a weighting vector (as the joint space may be noncommensurate if both prismatic and revolute joints occur), $\varepsilon$ is a step-width between the generated targets, and $t$ indicates the time-step. With a probability of $p^{home}\approx0.01\%$, $\ve{q}^{home}$ is selected as a target at each step.
	
	GBSCHEME is then called for each generated target and the robot tries to explore along the desired direction until its actual direction of motion deviates from the intended one more than $\varphi$ degrees. For $\varphi=\frac{\pi}{2}$, Eq.~(\ref{e2}) holds (cf.~l.~7 in Alg.~\ref{algo_ds}):
	
	\begin{equation}\label{e2}
	{\alpha=({\ve{q}^*_t} - {\ve{q}^*_{t-1}})}^T({\ve{q}^+_t - \ve{q}^+_{t-1}})
	\end{equation}
	
	where $\ve{q}^+_t$ is the currently observed configuration and $\ve{q}^+_{t-1}$ is the previous one. In this case, the agent will return to its previous configuration $\ve{q}^+_{t-1}$ to avoid drifting. Then, it will start following a new randomly selected direction again~\cite{GB,Rolf2013_DS}. 

\begin{algorithm}[!tb]
	\caption{Modified Direction Sampling}\label{algo_ds}
	\begin{algorithmic}[1]
		\Procedure {DS}{ $\ve{\tau}^{home}$, $\boldsymbol{SST}$}
		\State initialize learner $\ve{\hat{G}}(\ve{q}^{*}) = \ve{\tau}^{home}$ 
		\State choose random direction $\ve{\varDelta q}$
		\State \textbf{for $N$ number of samples} 
		\State \indent  generate target $\ve{q}^*_t$
		\State \indent \Call{GBScheme}{$\ve{q}_t^*$}
		\State \indent \textbf{if} $\alpha < 0$
		\State \indent \indent go to $\ve{q}^+_{t-1}$
		\State \indent \indent choose new direction $\ve{\varDelta q}$
		\State \indent \textbf{end if}
		\State \textbf{end for}
		\EndProcedure
	\end{algorithmic}
\end{algorithm}

\section{Accelerating Learning by Exploiting Symmetries}\label{sec:Sym}

The ISM generally associates each torque vector $\ve{\tau}$ with a non-singleton set $\mathcal{C}$ of configurations. Hence, relations between the configurations in $\mathcal{C}$ can be exploited to associate \emph{all} configurations in $\mathcal{C}$ with the applied torque vector $\ve{\tau}_\mathcal{C}$ by sampling just one configuration of $\mathcal{C}$. 

This idea is explored in the following paragraphs:
First, the notion of symmetries in the gravity term is introduced and exemplified. Subsequently, strategies to discover and exploit these symmetries to reduce the number of samples required for learning ISMs are presented. Finally, the achievable speedup and limitations of the current approach are briefly addressed.

\subsection{Symmetries in the Gravity Term}
The gravity term $\ve{G}(\ve{q})$ in the inverse dynamics model of a discretely-actuated serial manipulator constitutes a \emph{many-to-one} mapping between the set of admissible configurations $\mathcal{Q}_{\nu}$ and the set of required torques $\mathcal{T}_{s}$, i.e. $\ve{G}:\mathcal{Q}_{\nu}\rightarrow\mathcal{T}_{s}$ typically associates a member of the range $\mathcal{T}_{s}$ with more than one member of the domain $\mathcal{Q}_{\nu}$.
Therefore, there typically exist level sets 
\begin{equation}
\mathcal{L}_{\ve{\tau}}=\left\{ \ve{q} : \ve{G}(\ve{q})=\ve{\tau} \right\}
\end{equation}
with cardinalities $|\mathcal{L}_{\ve{\tau}}|>1$ for admissible torque vectors $\ve{\tau}$.

Two classes of configurations in these level sets can be distinguished.
Consider two level sets $\mathcal{L}_{\ve{\tau}_i}$ and $\mathcal{L}_{\ve{\tau}_j}$ where
\begin{eqnarray}
\label{eq:lvl}
\ve{\tau}_i=\ma{\varUpsilon}\ve{\tau}_j, \ma{\varUpsilon}=\diag\left(\delta_1,\hdots,\delta_n \right), 
\\\delta_\bullet=\pm1, \sum_{i=1}^{n}\delta_i<n\nonumber
\end{eqnarray}
i.e. the elements in $\ve{\tau}_i$ and $\ve{\tau}_j$ differ w.r.t. their sign. Here, $n$ denotes the number of joint DoFs and $\diag\left(\delta_1,\hdots,\delta_n \right)$ indicates a diagonal matrix with $\delta_1,\hdots,\delta_n$ on its main diagonal.

Primary symmetric configurations, also denoted as primary symmetries, constitute those configurations $\ve{q}_r,\ve{q}_s \in \bigcup_{k=1}^{2^n}\mathcal{L}_{\ve{\tau}_k}=\breve{\mathcal{L}}_{\ve{\tau}}$ in the union of all level sets fulfilling Eq.~(\ref{eq:lvl}) for which
 \begin{equation}
 \label{eq:prim_sym}
\ma{M}_{r,s}\ve{q}_r+\ma{N}_{r,s}\ve{q}_s=\ve{d}_{r,s}
 \end{equation}
 holds --
 where $\ve{d}_{r,s} \in \mathbb{R}^n$ and $\ma{M}_{r,s},\ma{N}_{r,s} \in \mathbb{R}^{n\times n}$ are constant (in particular independent of the choice of $\ve{\tau}$). 
 The set of all configurations in $\bigcup_{k=1}^{2^n}\mathcal{L}_{\ve{\tau}_k}$ which are directly or transitively related by Eq.~(\ref{eq:prim_sym}) is called the set of primary symmetries~(SPS) denoted by $\mathcal{S}$. 
 
 Secondary symmetric configurations constitute those configurations for which at least one of $\ve{d}_{r,s},\ve{M}_{r,s}, \ve{N}_{r,s} $ is a function of $\ve{q}$ and/or $\ve{\tau}$.

Clearly, the relations between primary symmetries can be inferred quite easily given a number of level sets, e.g. by multiple linear regression. In contrast, the relations between secondary symmetries are far harder to deduce.
Therefore, in the experimental results, we will only consider primary symmetries.

\begin{figure}[!tb]
	\centering
	\includegraphics[width=0.4\linewidth]{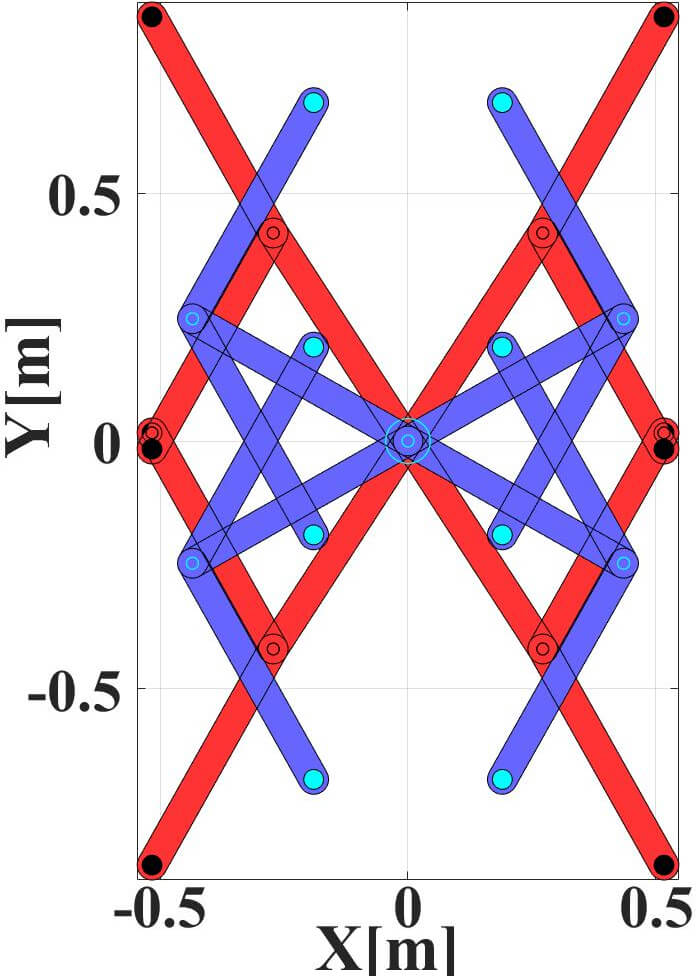}
	\caption{Symmetric configurations of a 2R planar robot. Configuration pairs in the configuration sets illustrated in red  $\mathcal{S}_B$ (and blue  $\mathcal{S}_A$ resp.) are primary symmetric to each other. Configurations of either set are related to configurations in the other set by secondary symmetries. Note that the manipulator is stretched out to the right in its zero configuration and that the gravity vector points downwards into negative y-direction.}
	\label{fig:sym_arm}
\end{figure}

\begin{figure}[!tb]
	\centering
	\includegraphics[width=0.9\linewidth]{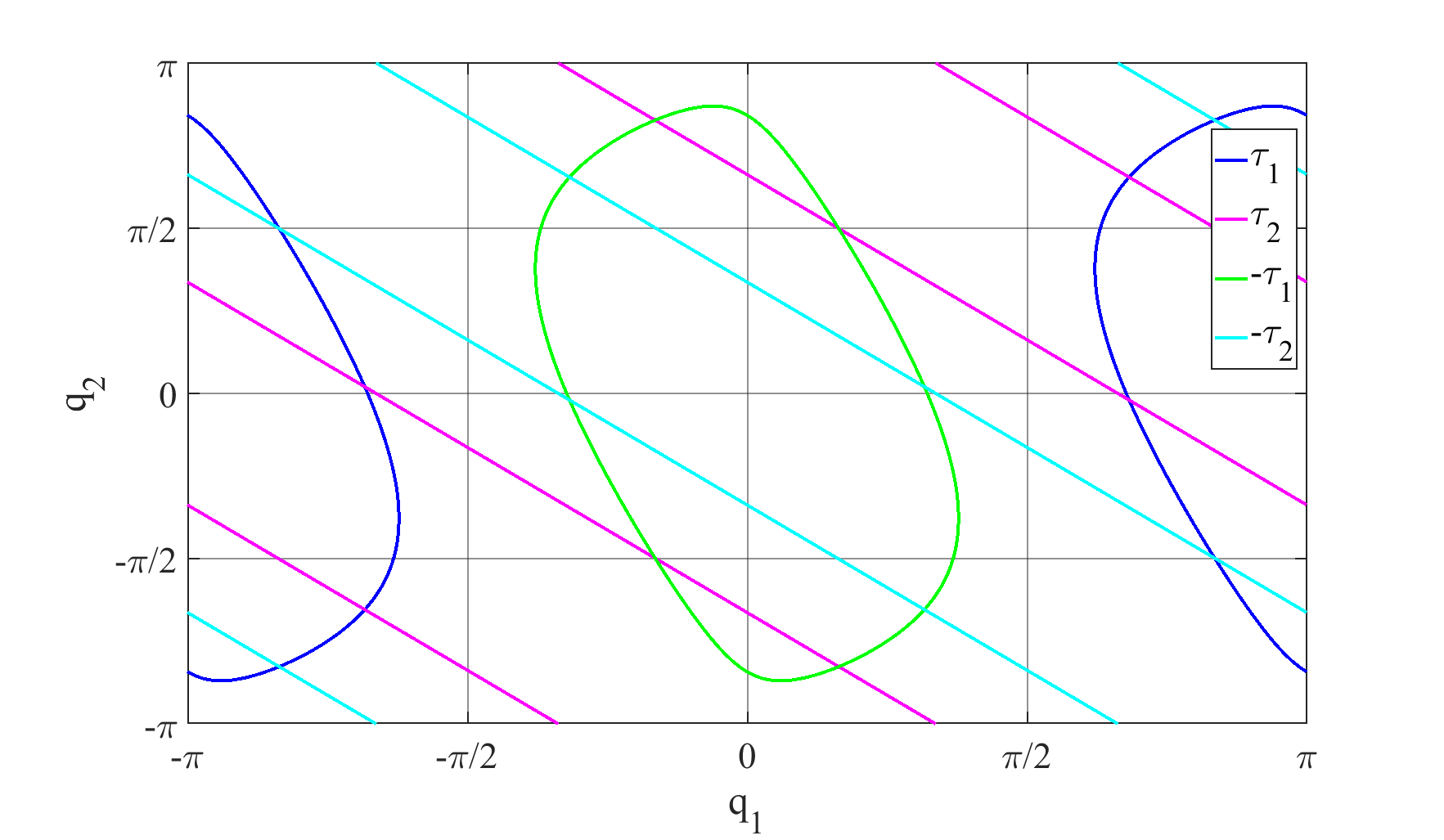}
	\caption{Torque level sets of a 2R planar manipulator corresponding to the configurations in Fig.~\ref{fig:sym_arm}. The 16 intersection points constitute symmetric configurations. Their color corresponds to Fig.~\ref{fig:sym_arm}.}
	\label{fig:symmetries}
\end{figure}

To exemplify the idea of primary and secondary symmetries, Fig.~\ref{fig:sym_arm} shows all symmetric configurations of a 2R planar robot. There are 16 configurations which can be separated into two disjoint sets $\mathcal{S}_A$ (blue) and $\mathcal{S}_B$ (red) of $8$ configurations each. Each set $\mathcal{S}_A$ and $\mathcal{S}_B$ constitutes a set of primary symmetries; secondary symmetries occur by relating configurations from $\mathcal{S}_A$ with those from $\mathcal{S}_B$.
Fig.~\ref{fig:symmetries} shows the corresponding level sets of the 2R planar manipulator for the configurations depicted in Fig.~\ref{fig:sym_arm}. All intersection points of level sets for $\pm \tau_1$ and $\pm \tau_2$ constitute symmetric configurations as they possess the same absolute value of the elements in the torque vector. As in Fig.~\ref{fig:sym_arm}, the $16$ intersection points in Fig.~\ref{fig:symmetries} can be separated into the two disjoint sets $\mathcal{S}_A$ and $\mathcal{S}_B$ indicated by the color of the points.
\subsection{Utilizing Symmetric Configurations}
\label{sec:acc_learn_sym}
Symmetric configurations can be discovered by applying suitable torque profiles to the manipulator. Subsequently, relations between different $\ve{q}_i$ in level-sets $\breve{\mathcal{L}}_{\ve{\tau}}$ can be established. Once a number of $n_{sym}$ functional relations is determined, each applied motor command $\ve{\tau}_i$ generates a sample $(\ve{q}_i,\ve{\tau}_i)$ as well as $n_{sym}-1$ further samples $(\ve{q}_j,\ve{\tau}_i), i\neq j, j \in {1,\hdots,n_{sym}-1}$ obtained by evaluating the previously established functional relations between symmetric configurations. The details of this process are addressed in the following subsections.

\subsubsection{Discovering Symmetric Configurations}
\begin{figure}
	\centering
	\includegraphics[width=1.0\linewidth]{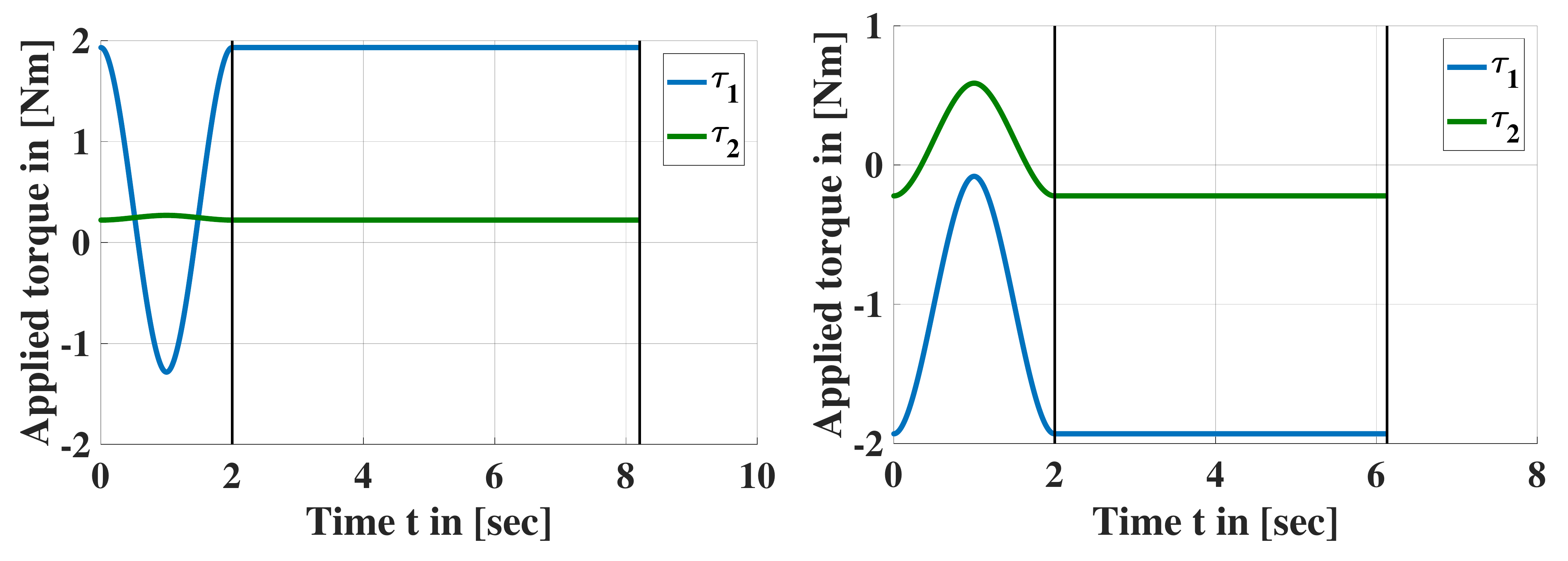}
	\caption{Examples of torque profiles for symmetry discovery. First, a torque spline is applied. Subsequently, a constant torque is commanded until the manipulator settles at a configuration.}
	\label{fig:torqueprofiles}
\end{figure}

\begin{figure}[htbp!]
	\centering
	\includegraphics[width=1.0\linewidth]{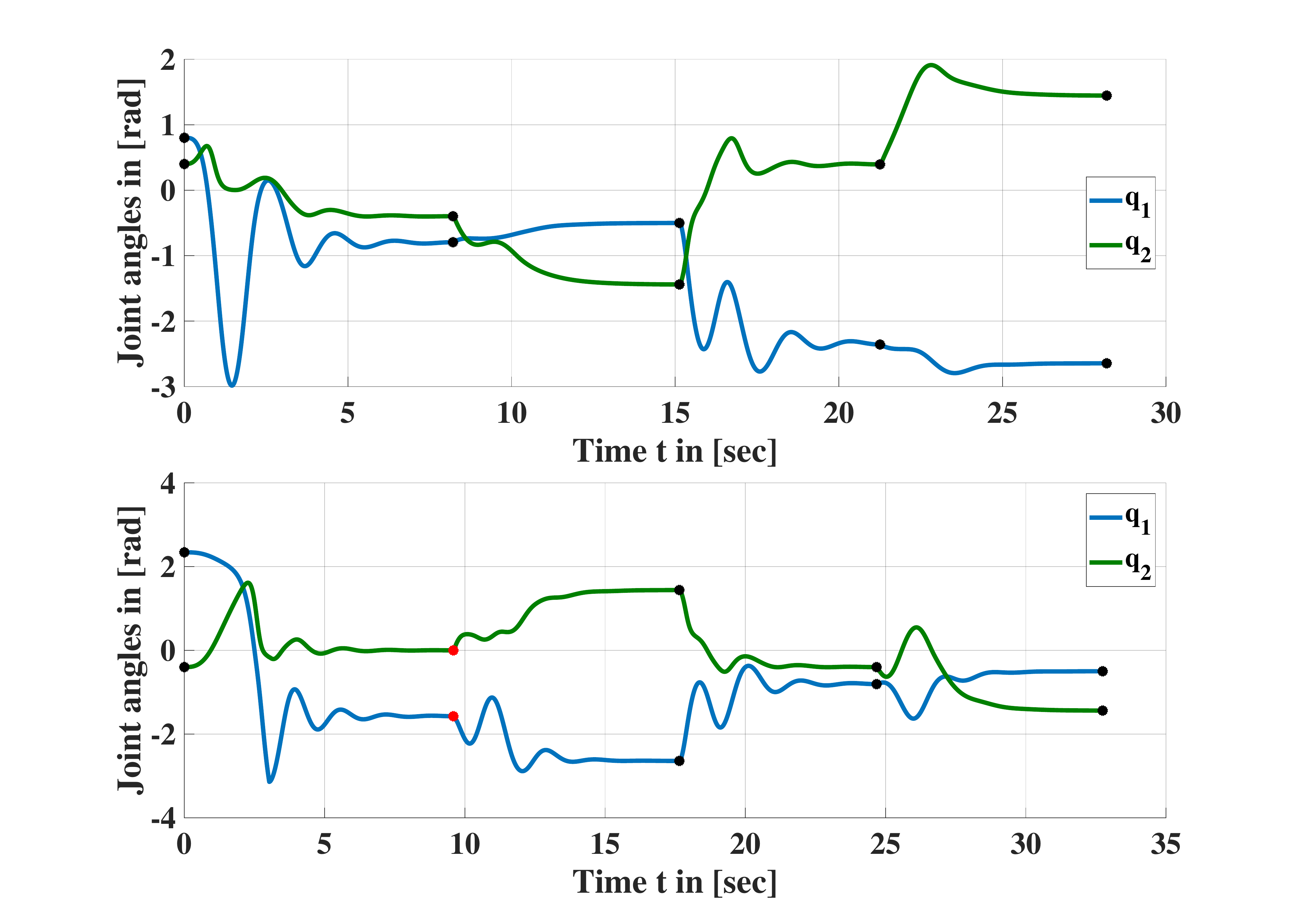}
	\caption{Joint trajectories resulting from the application of torque profiles according to Fig.~\ref{fig:torqueprofiles}. Red dots indicate that the joint limits have been reached and the manipulator returned to its home configuration. Black dots indicate the end of a profile. The corresponding configurations are entered into $\breve{\mathcal{L}}'_\tau$ (cf.~Alg.~\ref{algo:symmetry_discovery}).}.
	\label{fig:jointtrajectories}
\end{figure}

For symmetry discovery, sequences of suitable torque profiles are applied for numerous torque vectors.
Alg.~\ref{algo:symmetry_discovery} shows the required steps for discovering the symmetries associated with a single torque vector $\ve{\tau^*}$.
Starting from the home configuration, a sequence of $n_{pr}$ torque profiles $\ve{\tau}^p_i[k], k \in [1,..,n_{s_i}]$~(cf.~Fig.~\ref{fig:torqueprofiles}) of $n_{s_i}$ samples are generated using splines with start and end-point constraints on their derivatives, i.e. $\dot{\ve{\tau}}^p_i[1]=\dot{\ve{\tau}}^p_i[n_{s_i}]=\ve{0}$. Probability distributions $p_\tau$ and $p_n$ are utilized to draw $n_{s_i}$ and to generate intermediate torques in each profile.
For each $\ve{\tau}^p_i[k]$, $\ve{\tau}^p_i[1]=\ma{\varUpsilon}\ve{\tau}^* \wedge \ve{\tau}^p_i[n_{s_i}]=\ma{\varUpsilon}\ve{\tau^*}$ holds. By always reverting to the same torque magnitude at the end of each profile but applying different intermediate torques, different $\ve{q}$ -- related by primary or secondary symmetries -- are (likely to be) reached. If the manipulator reaches its joint limits during or after application of a torque profile, it goes back to its home configuration $\ve{q}^{home}$ again and the sequence is continued with the next profile.
After successful application of a torque profile, $\ve{\tau}_i^p[n_{s_i}]$ is applied as long as the manipulator has not settled.
If the manipulator settles at a valid configuration, this configuration is recorded and added to the discovered set $\breve{\mathcal{L}}'_{\tau}$ (if not already contained in it) associated with the torque $\ve{\tau}$.
\begin{algorithm}[!tb]
	\caption{Symmetry Discovery}\label{algo:symmetry_discovery}
	\begin{algorithmic}[1]
		\Function{discoverSym}{$\ve{\tau^*},\ve{q}_{min},\ve{q}_{max},n_{pr},p_n,p_\tau$}
		\State $\breve{\mathcal{L}}'_{\tau}=\emptyset$
		\State \textbf{for} {$i=1,\hdots,n_{pr}$}
			\State \indent $settled=false$
			\State \indent $n_{s_i}=$ {sample probability distribution $p_n$}
			\State \indent $\ve{\tau}_i^p=$\Call{generateTorqueProfile}{$\ve{\tau^*},p_\tau,n_{s_i}$}
			\State \indent \textbf{for} {$j=1,\hdots,n_{s_i}$}
				\State \indent \indent apply torque {$\ve{\tau}_i^p[j]$}
				\State \indent \indent observe current configuration $\ve{q}_{curr}$
				\State \indent \indent \textbf{if}  {$\ve{q}_{curr}$ exceeds joint limits}  
				\State \indent \indent \indent go home
				\State \indent \indent \indent  $settled=true$
				\State \indent \indent \textbf {end if}				
			\State \indent \textbf {end for}	
			\State \indent \textbf {while}{$\neg settled$}
				\State \indent \indent apply torque $\ve{\tau}^p_i[{n_{s_i}}]$
				\State \indent \indent observe current configuration $\ve{q}_{curr}$
				\State \indent \indent \textbf{if}  {$\ve{q}_{curr}$ exceeds joint limits}  
				\State \indent \indent \indent go home
				\State \indent \indent \indent $settled=true$
				\State \indent \indent \textbf {elseif} $\ve{q_{curr}}$ has settled
				\State \indent \indent \indent add $\ve{q}_{curr}$ to \textbf {${\mathcal{L}}'_{\tau}$}			
				\State \indent \indent \textbf {end if}			
		    	\State \indent  \textbf {end while}	
		\State \textbf {end for}			
		\State\Return $\breve{\mathcal{L}}'_{\tau}$
		\EndFunction
	\end{algorithmic}
\end{algorithm}
Fig.\ref{fig:torqueprofiles} shows exemplary torque profiles and Fig.\ref{fig:jointtrajectories} shows two joint trajectories resulting from the application of such torque profiles which discover $9$ symmetric configurations.
\subsubsection{Learning Functional Relations and Exploiting Symmetric Configurations}
Functional relations between primary symmetries according to Eq.~\ref{eq:prim_sym} can be easily determined by established multilinear linear regression techniques~(cf. e.g.~\cite{Draper:98}). These learned relations can then be utilized to compute symmetric configurations for each observed $\ve{q}$ and assign the same torque $\ve{\tau}$ to all of them. 


\subsubsection{Integration into ISM Learning Schemes}
\begin{figure}[htbp!]
	\centering
	\includegraphics[width=0.95\linewidth]{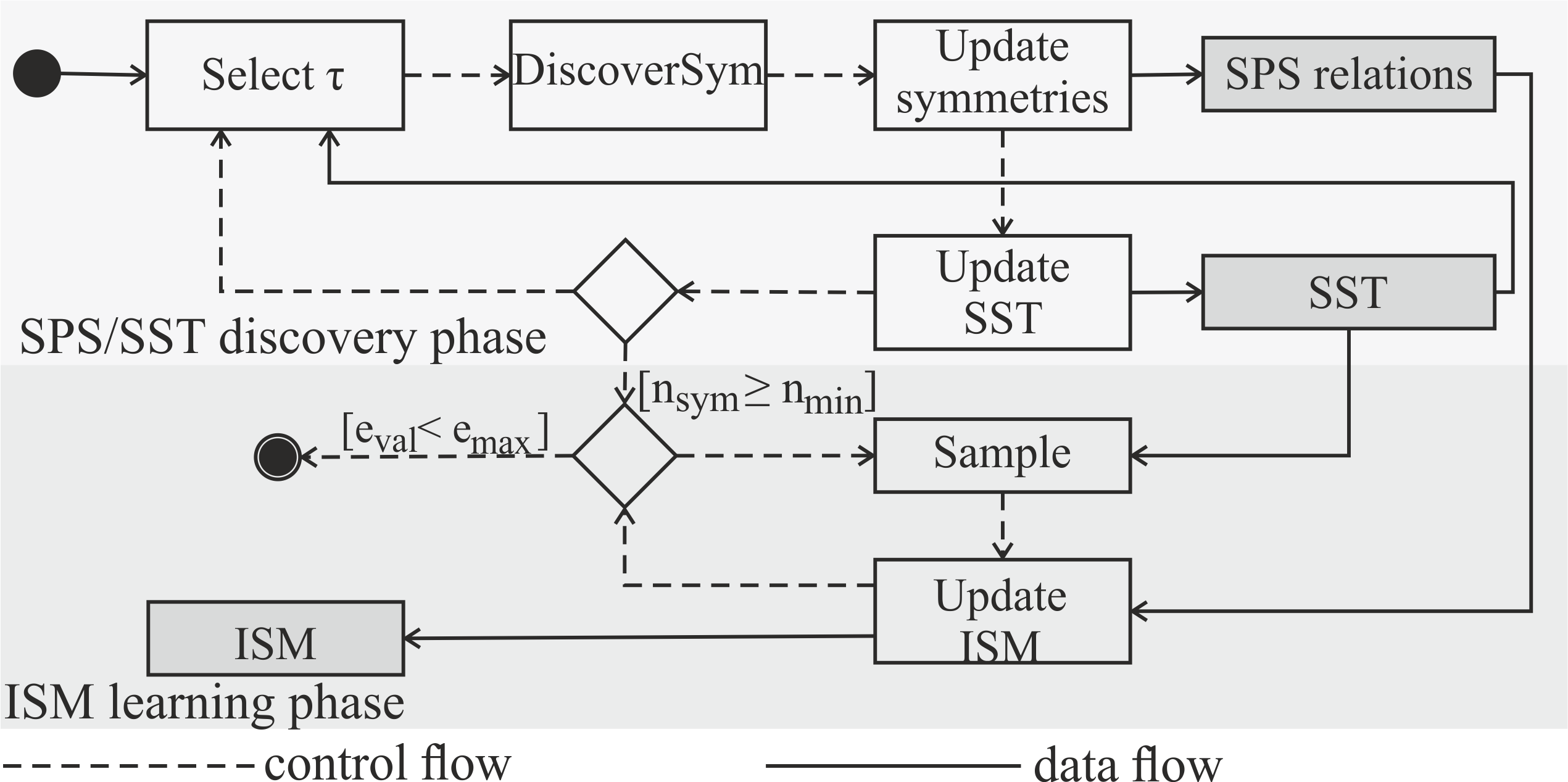}
	\caption{Illustration of the SST and SPS discovery as well as the ISM learning phase. The estimated SST is used to generate admissible torque samples and the SPS is used to generate $n_{sym}$ training samples from $1$ recorded sample.}
	\label{fig:symmetries_flow_chart}
\end{figure}

Symmetries in ISMs can be exploited with almost any learning approach.
Fig.~\ref{fig:symmetries_flow_chart} illustrates the required steps for symmetry discovery and exploitation.
In the discovery phase, a target torque $\ve{\tau}$ is selected. Subsequently, Alg.~\ref{algo:symmetry_discovery} is applied to discover symmetric configurations. Then, multiple linear regression is performed using the output of Alg.~\ref{algo:symmetry_discovery} to update the relations between primary symmetries. The applied torque profiles and observed joint angles are exploited to update the estimates of the $\boldsymbol{SST}$ and optionally the $\boldsymbol{BCTS}$~(cf. Sec.~\ref{sec:BCTS}). When a sufficient number $n_{sym}\geq n_{min}$ of symmetries has been discovered, the learning phase begins and the functional relations between the primary symmetries are exploited to generate $n_{sym}$ training samples based on one applied training torque vector. Direction Sampling (cf.~Alg.~\ref{algo_ds}) or any other online (or batch) learning approach can be applied to obtain the ISM. The learning phase is terminated if a desired validation error $e_{val}$ is reached.

\subsection{Achievable Speed-Up}
For the planar 2R robot addressed in Sec.~\ref{2D}, the currently achievable reduction factor $r$ w.r.t. required samples is limited by $r=8$ as the primary set of symmetries has cardinality $8$, while exploiting secondary symmetries would further increase $r$. For the 3R manipulator addressed in Sec.~\ref{3D}, the cardinality of the primary set of symmetries and $r$ increase to $r=16$. Exploiting secondary symmetries would again yield far higher reduction factors depending on the properties and configuration of the manipulator.


\subsection{Limitations}
Currently, our approach is limited to primary symmetries as efficient learning of the functional relations between secondary symmetries proves to be challenging.
Furthermore, elasticity, gear effects as well as nonlinear friction effects are currently not considered but may affect the discovered symmetries in experiments with geared, elastic manipulators.

\section{Experimental Results}\label{sec:expermints}
This section presents various results for learning ISMs with modified online Goal Babbling and shows the efficiency gains resulting from symmetry exploitation with both online and batch learning techniques.

\subsection{Learning ISMs with Modified Online Goal Babbling:}\label{gb_result}
\begin{figure}[!tb]
	\centering
	\subfigure[RTB model for a 2R planar robot.]{\label{2D} \includegraphics[height=50mm,width=0.4\textwidth]{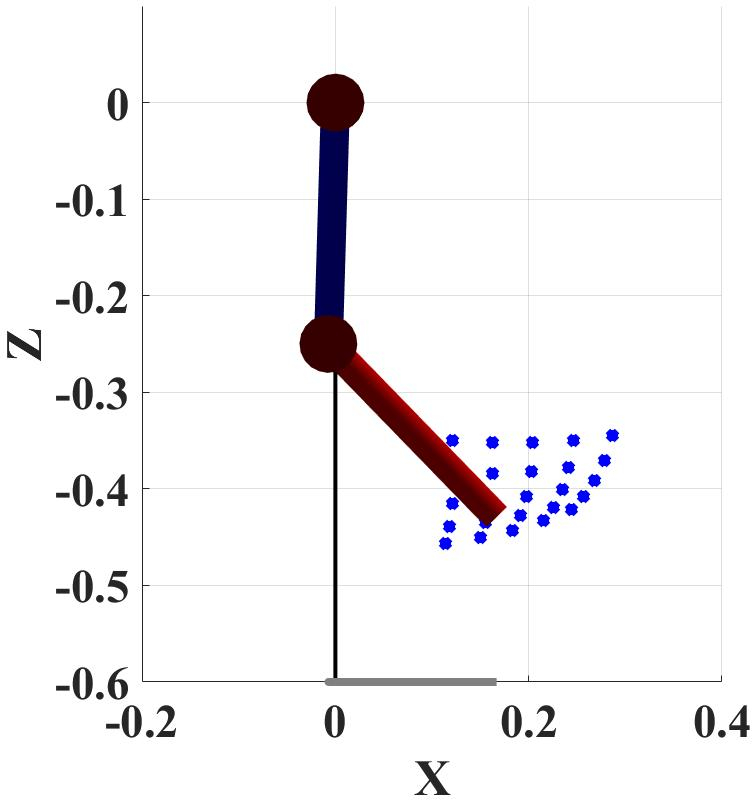}}\hspace{1mm}
	\subfigure[3R simplified human arm.]{\label{3D} \includegraphics[height=50mm,width=0.4\textwidth]{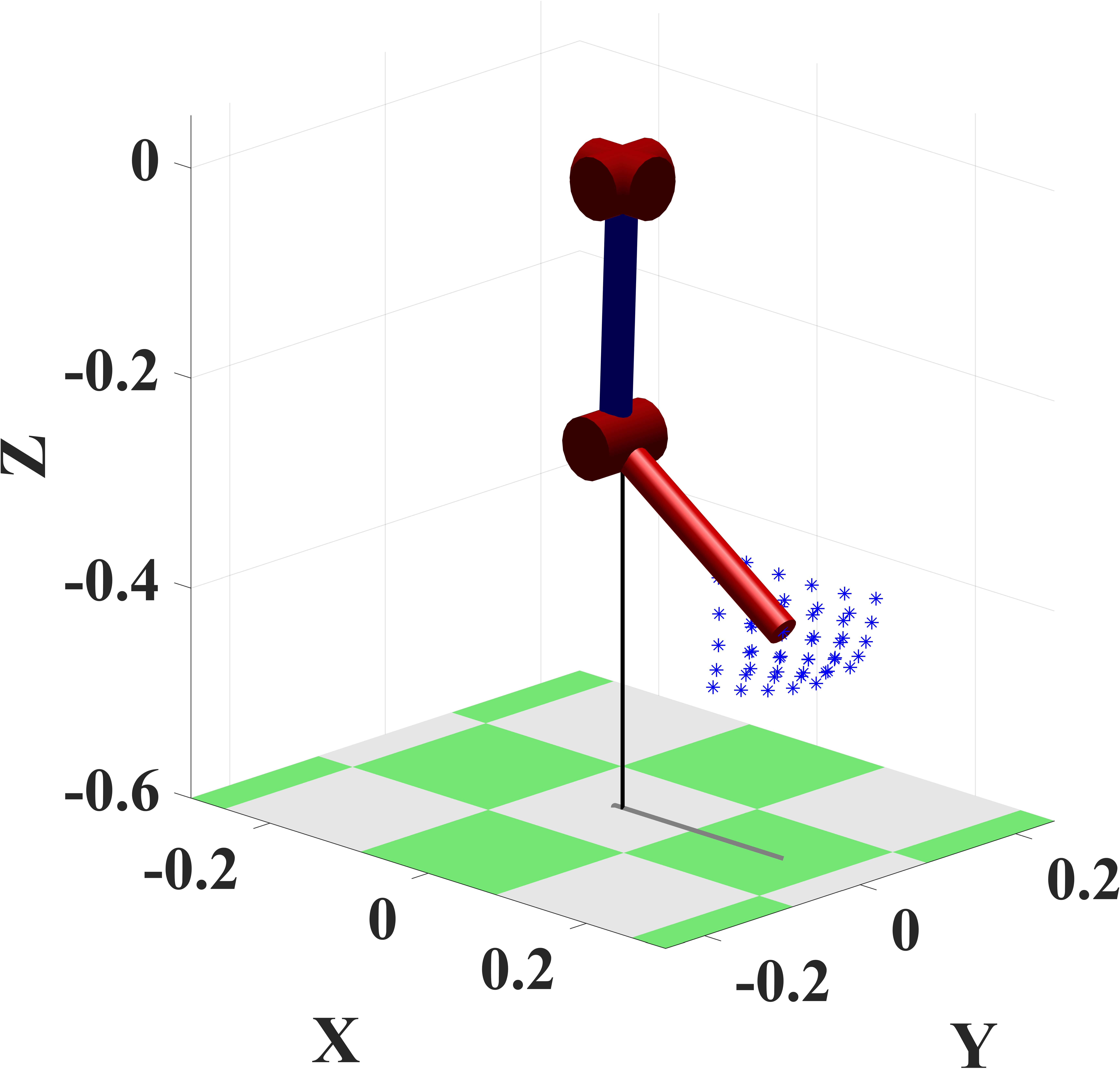}}
	\subfigure[4R humanoid arm]{\label{4D}
    \includegraphics[height=50mm,width=0.4\textwidth]{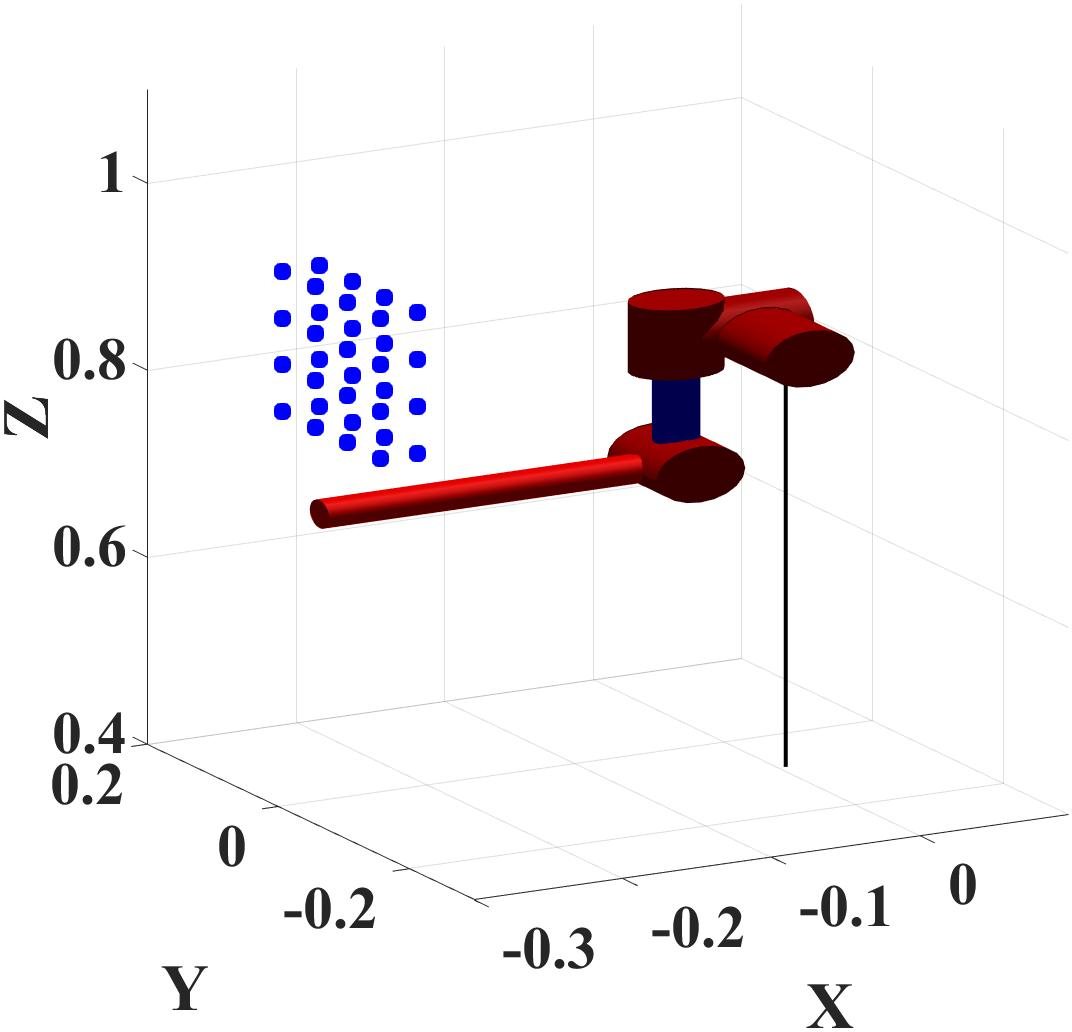}}\hspace{1mm} 
\subfigure[Training Error for 4R arm.]{\label{err} \includegraphics[height=50mm,width=0.4\textwidth]{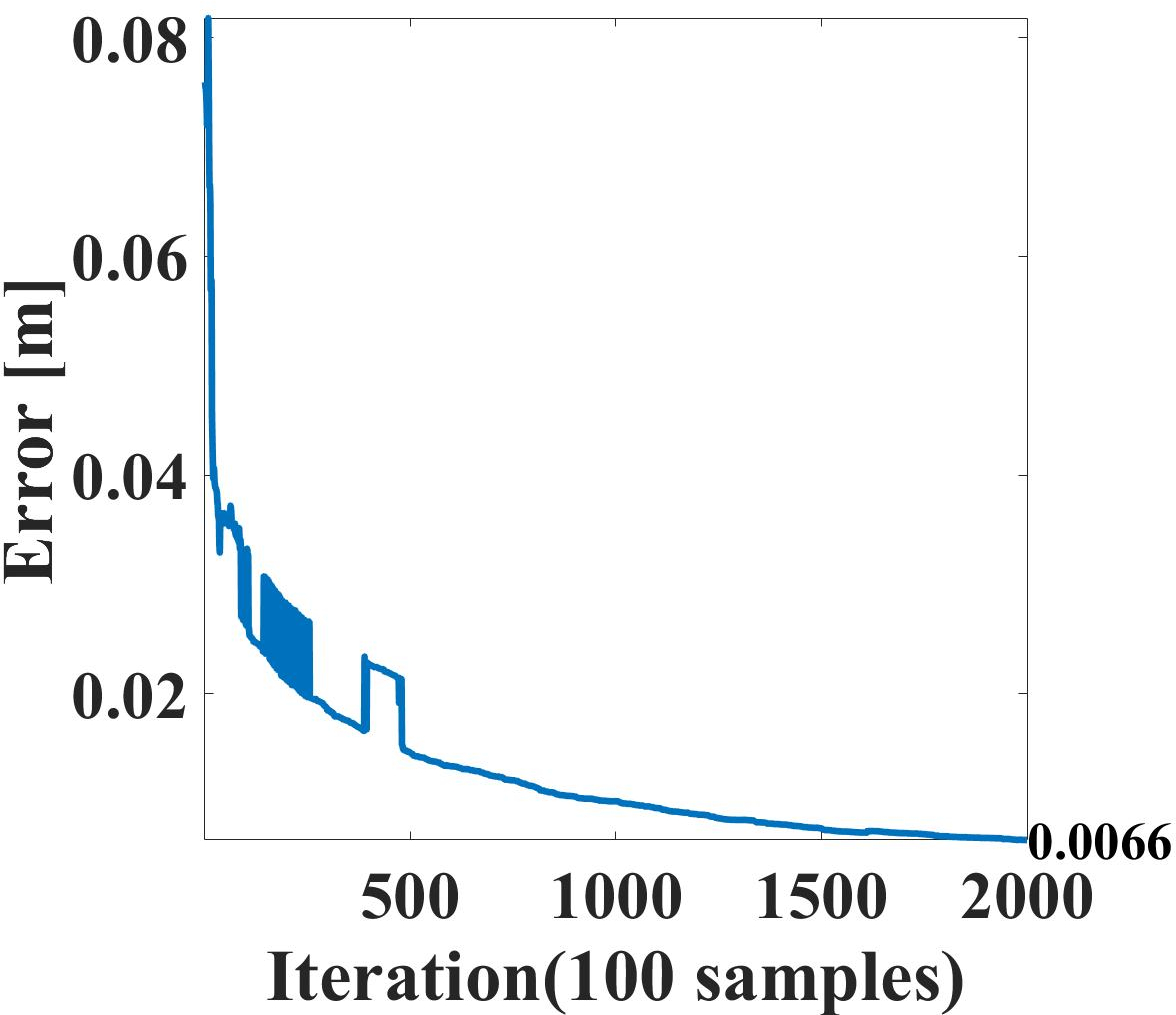}}\hspace{1mm} 
	\caption{(a), (b), and (c): Robots with targets. (d): Training error for 4R humanoid arm. }
	\label{fig:Robots}
\end{figure}

We present experimental results for a 2R planar robot, a 3R simplified human arm~\cite{Arm} and a 4R humanoid arm~\cite{COMAN}. The models have been set up in MATLAB using the Robotic Toolbox (RTB)~\cite{Robotic} and the targets are distributed in task space as illustrated in Fig.~\ref{fig:Robots}. However, in the online Goal Babbling scheme for learning the ISMs, the corresponding joint configurations are employed.

\begin{figure}[!tb]
	\centering
	\subfigure[Discovered $\boldsymbol{SST}$.]{\label{tau2} \includegraphics[height=50mm,width=0.4\textwidth]{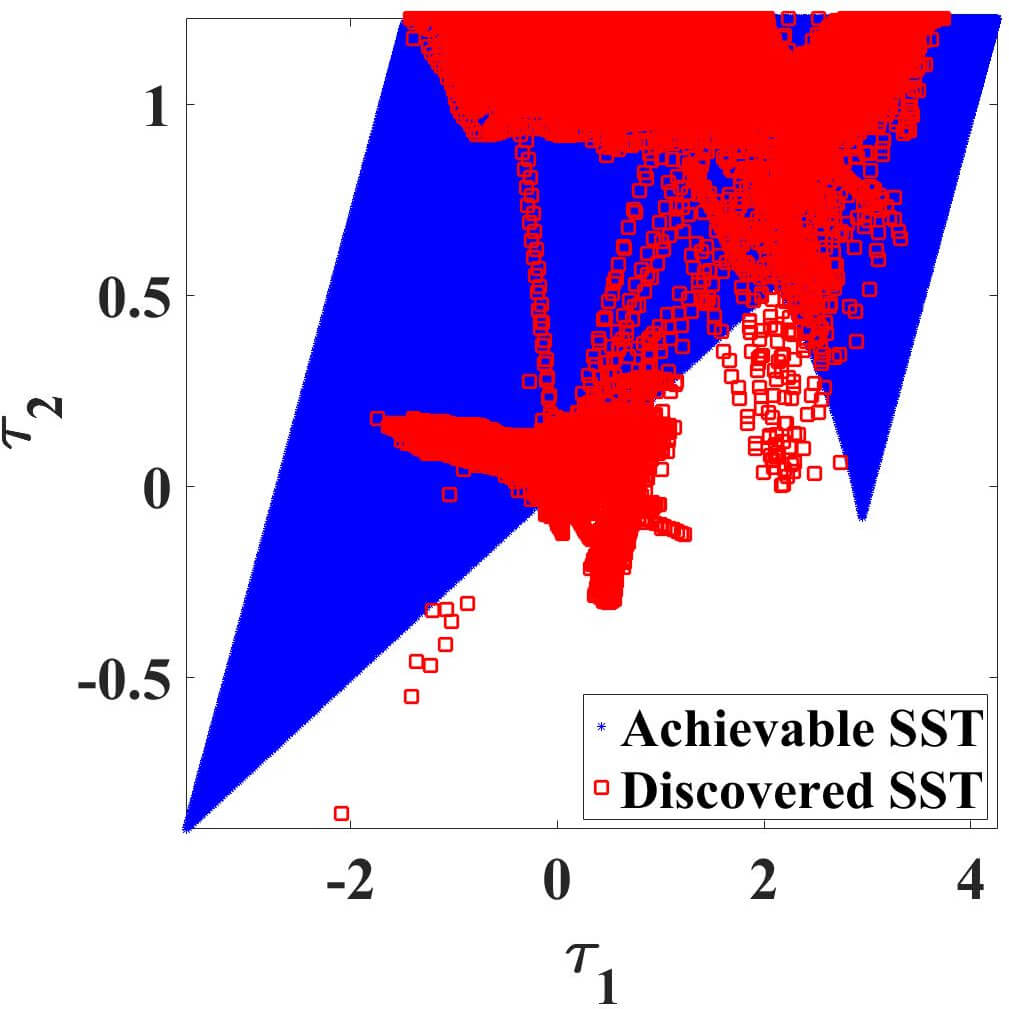}}
	\subfigure[Corrected discovered $\boldsymbol{SST}$.]{\label{SOST} \includegraphics[height=50mm,width=0.4\textwidth]{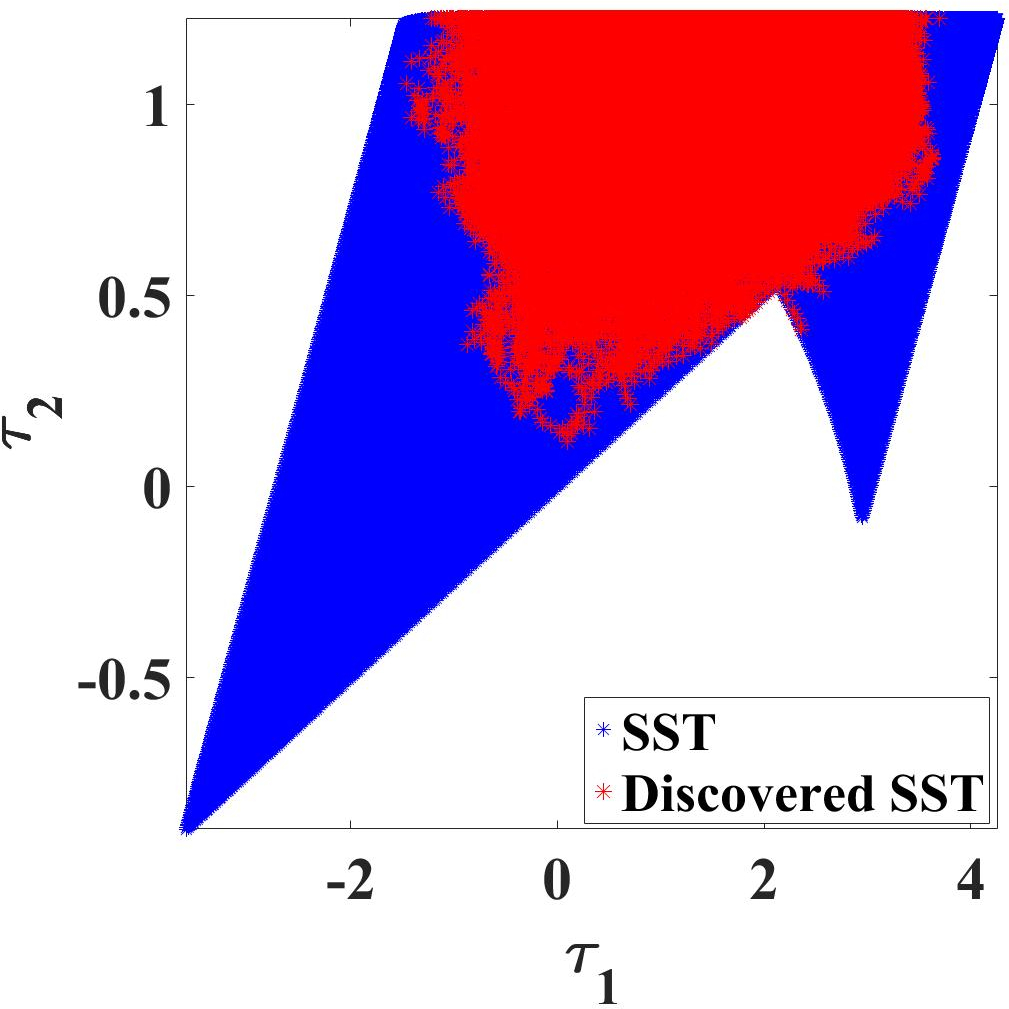}}
	\caption{$\boldsymbol{SST}$ of the 2R planar robot and discovered torques.}
	\label{fig:Tau}
\end{figure}

First, the $\boldsymbol{SST}$ was estimated for each robot~(cf.~Sec.~\ref{sec:SOST}); Fig.~\ref{fig:Tau} represents the estimated $\boldsymbol{SST}$~(blue area) for the 2R planar robot with specific joint limits. Fig.~\ref{tau2} illustrates that applied torques may not be contained in the $\boldsymbol{SST}$~(red dots outside the $\boldsymbol{SST}$) due to exploratory noise. 

By applying the nearest neighbor strategy (cf. Sec.~\ref{sec:SOST}), the $\boldsymbol{SST}$ was correctly explored as shown in Fig.~\ref{SOST}.

Based on the $\boldsymbol{SST}$ estimate, online Goal Babbling according to Alg.~\ref{algo_gb} was applied.
It shows very good accuracies even for the 4R humanoid arm with a training root-mean-square error (RMSE) of less than $7mm$ in task-space as illustrated in Fig.~\ref{err}. 

\begin{figure}[!tb]
	\centering
	\subfigure[Training and testing targets]{\label{targets} \includegraphics[height=50mm,width=0.4\textwidth]{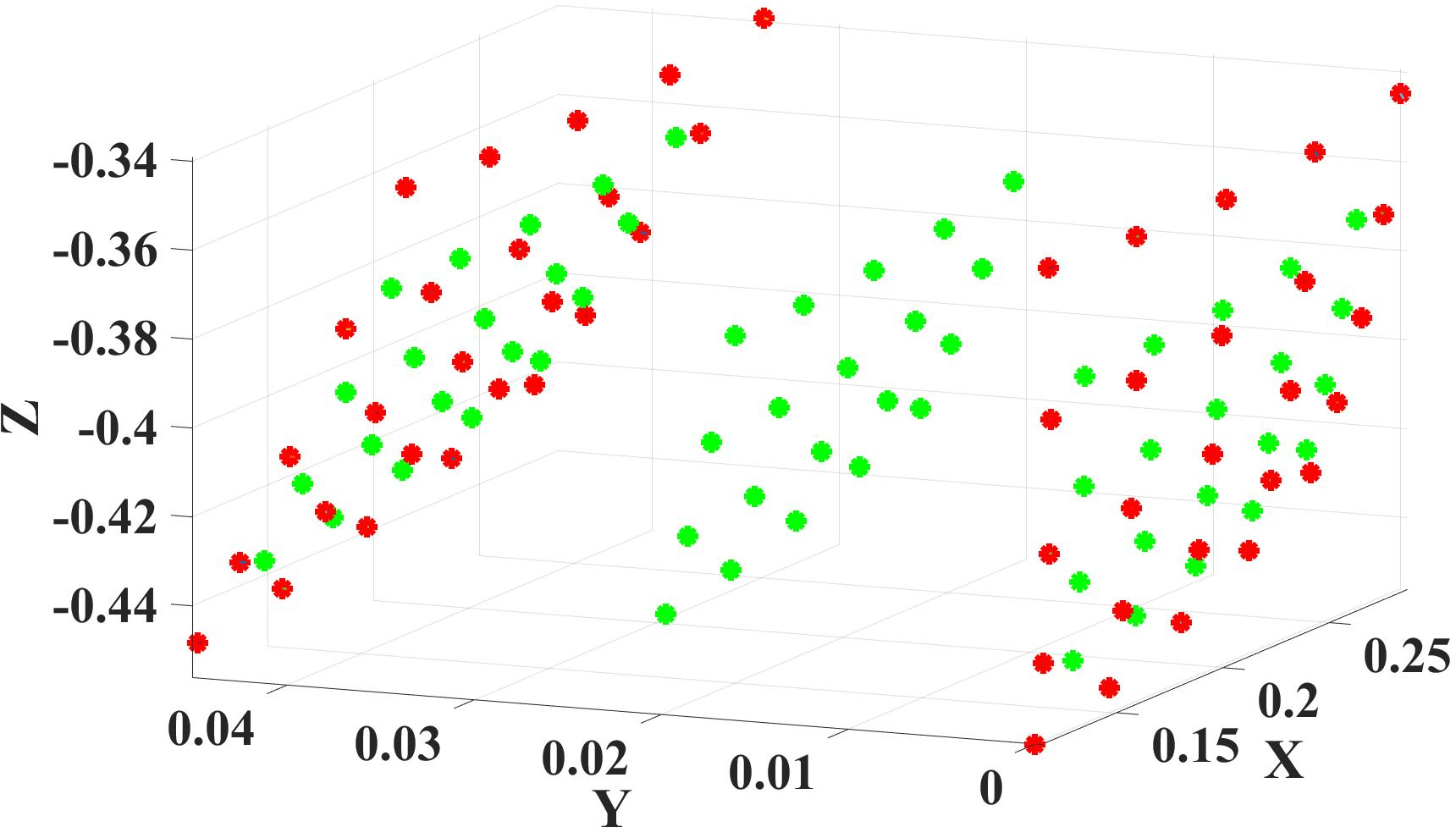}}\hspace{1mm}
	\subfigure[Training and testing results ]{\label{4D_targets1} \includegraphics[height=50mm,width=0.4\textwidth]{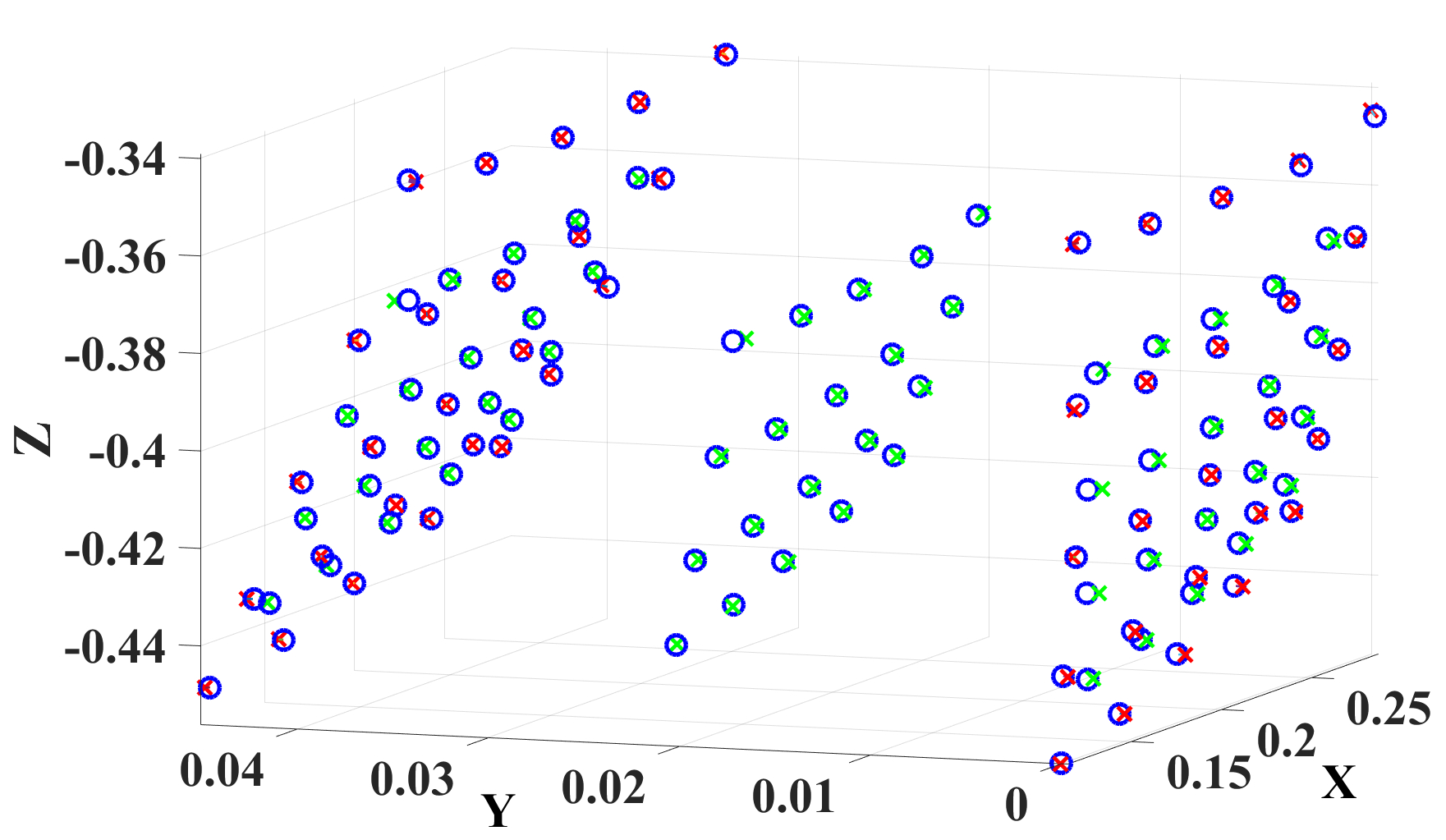}}
	\caption{Targets and testing results: red dots denote training targets, green dots testing targets, and blue dots denote the observed end-effector positions. Numerical results are provided in row 5 of Tbl.~\ref{tab:exp_results}.}
	\label{fig:Targets}
\end{figure}

The 2R, 3R, and 4R robots were trained to learn the required torque necessary to achieve their targets and maintain the desired configurations. 
Generalization was checked by defining suitable testing targets.
In Fig.~\ref{targets}, the training (red dots) and testing (green dots) targets for the 3R arm are visualized. The robot performed very well for both the training and testing sets as depicted in Fig.~\ref{4D_targets1}; all configuration targets were maintained with an RMSE of $0.5mm$ for the testing targets and $0.6mm$ for the training targets.  

Tbl.~\ref{tab:exp_results} shows various experimental performance results for the three employed robot models. The results demonstrate that the achieved accuracy is more than sufficient for most handling tasks.

\begin{table}[!tb] \label{T1}
	\caption{Experimental Results for Different Robots with modified online Goal Babbling}
	\label{tab:exp_results}
	\begin{center}
	\footnotesize
		\begin{tabular}{ |l|c|c|c|c|c| }
			\hline		
			\multirow{0}{2em}{Robot} &{Grid} & Nr of& Training& Nr of & Testing \\
		
			Model & Size $[cm]$ & Targets & Error $[m]$ & Samples & Error $[m]$\\

			\hline
			2R& $6\!\cdot\!3$  &  $25$ & $10^{\scalebox{0.40}[1.0]{\( - \)}4}$ & $6\!\cdot\!10^4$ &$10^{\scalebox{0.40}[1.0]{\( - \)}4}$\\
				
			2R& $15\!\cdot\!12$ & $25$  & $8\!\cdot\!10^{\scalebox{0.40}[1.0]{\( - \)}4}$ &$8\!\cdot\!10^4$&$5\!\cdot\!10^{\scalebox{0.40}[1.0]{\( - \)}4}$\\
			
			3R& $17\!\cdot\!11$&  $25$  &$4\!\cdot\!10^{\scalebox{0.40}[1.0]{\( - \)}4}$ & $3\!\cdot\!10^5$ & $3\!\cdot\!10^{\scalebox{0.40}[1.0]{\( - \)}4}$ \\
			
			3R& $6\!\cdot\!5\!\cdot\!1,8$& $60 $  & $5\!\cdot\!10^{\scalebox{0.40}[1.0]{\( - \)}4}$ & $3\!\cdot\!10^5$ & $5\!\cdot\!10^{\scalebox{0.40}[1.0]{\( - \)}4}$\\
			
			3R&   $17\!\cdot\!4\!\cdot\!11$ &   50 & $5\!\cdot\!10^{\scalebox{0.40}[1.0]{\( - \)}4}$ &$3\!\cdot\!10^5$ & $6\!\cdot\!10^{\scalebox{0.40}[1.0]{\( - \)}4}$ \\
			
			4R&   $16^2\!\cdot\!2,5$  & $32$ &$6\!\cdot\!10^{\scalebox{0.40}[1.0]{\( - \)}3}$ & $2\!\cdot\!10^5$& $4\!\cdot\!10^{\scalebox{0.40}[1.0]{\( - \)}3}$ \\
		
			\hline
		\end{tabular}
	\end{center}
\end{table}

\subsection{Learning ISMs with Exploiting Symmetries}
Owing to the symmetric properties of ISMs, only a fraction of the configuration space, denoted as bijective configuration-torque set ($\boldsymbol{BCTS}$), needs to be explored. First, the notion of the $\boldsymbol{BCTS}$ is introduced.
In principle, any exploration strategy and any suitable online/offline learning technique could be used to exploit symmetries and learn the ISM. To demonstrate the general applicability of symmetries, we then present results for learning ISMs with modified Direction Sampling (online) and with a batch learning scheme.

\subsubsection{Bijective Configuration-Torque Set}
\label{sec:BCTS}
\begin{figure}[!tb]
	\centering
	\subfigure{\label{ActiveJ} \includegraphics[height=50mm,width=0.4\textwidth]{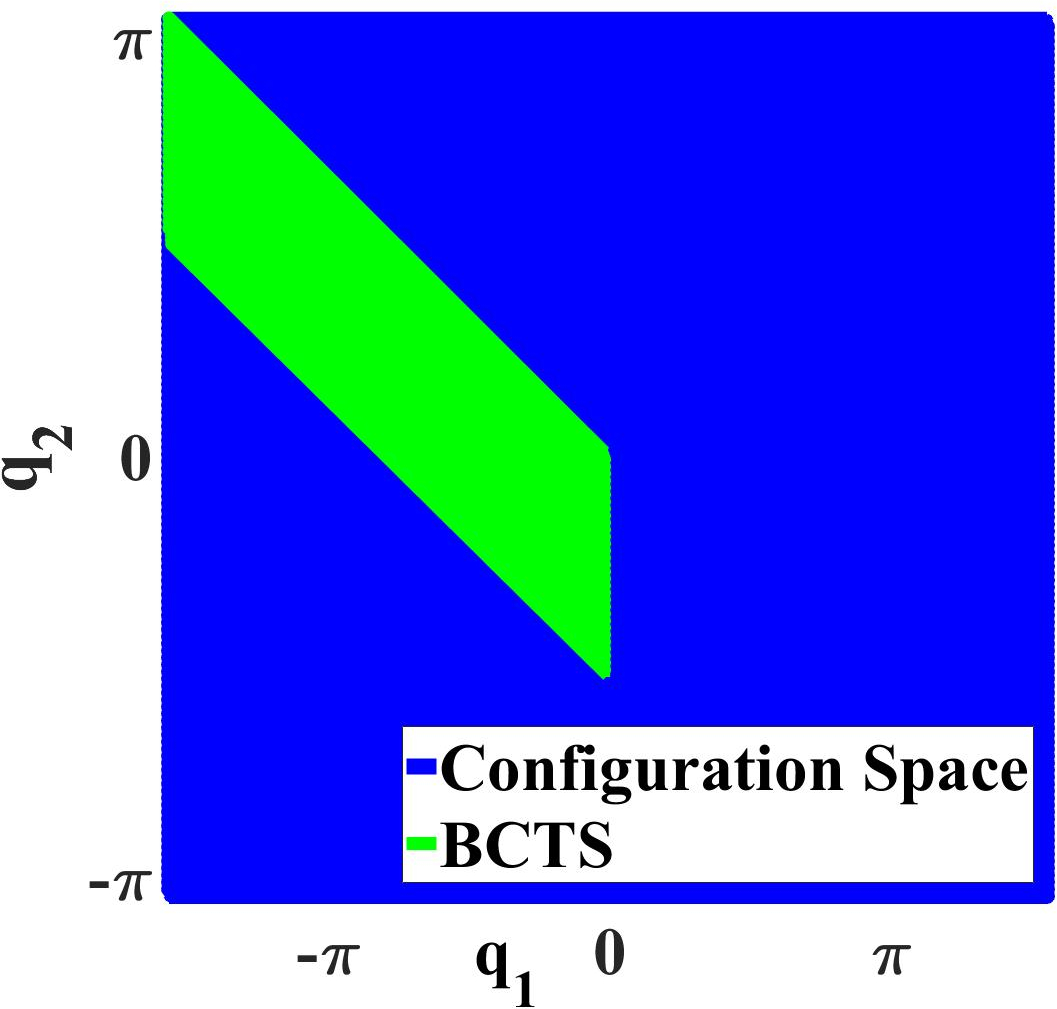}}
	\subfigure{\label{ActiveSOST} \includegraphics[height=50mm,width=0.4\textwidth]{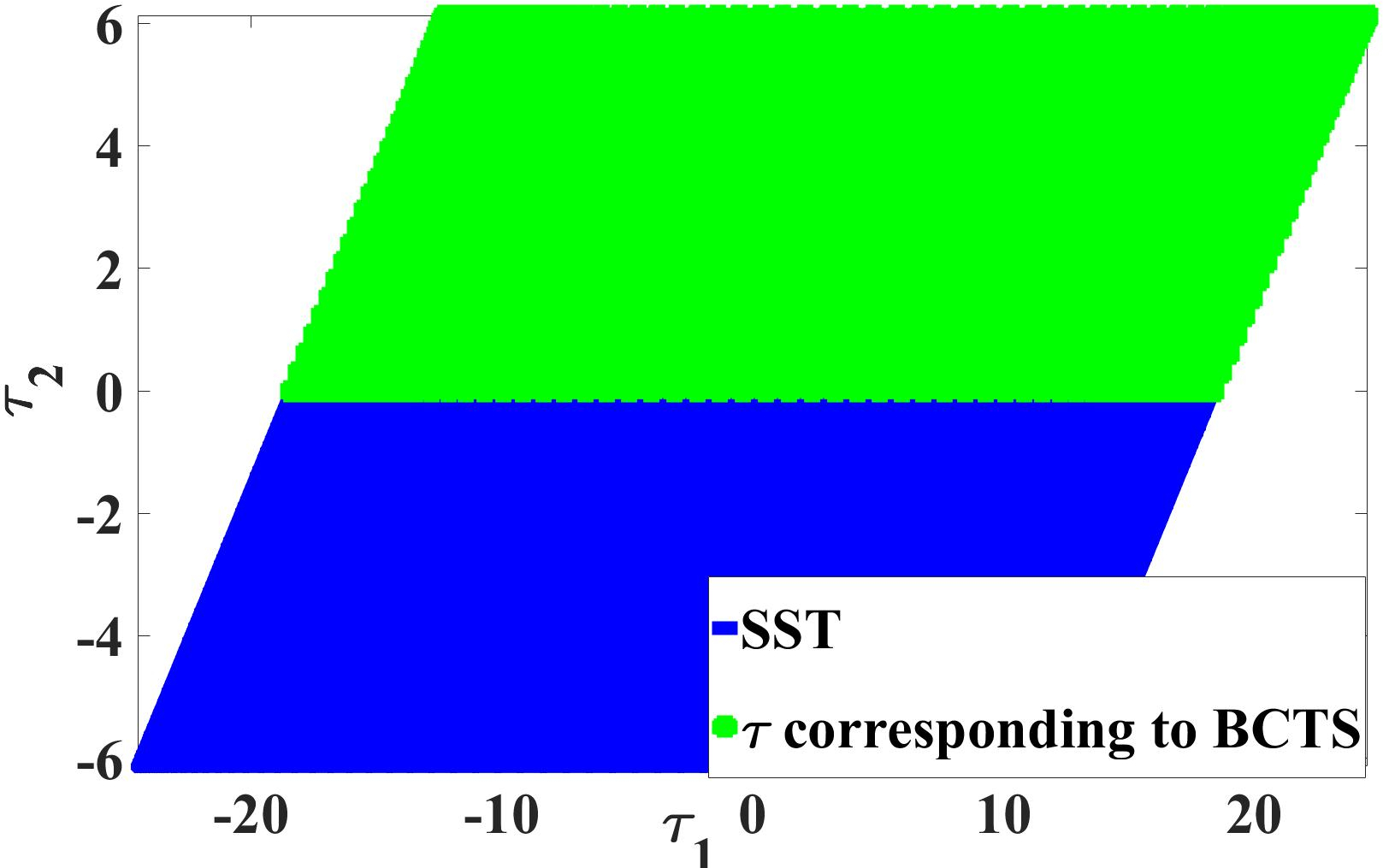}}
	\caption{(a): $\boldsymbol{BCTS}$ in configuration space; (b): torques set corresponding to $\boldsymbol{BCTS}$ in $\boldsymbol{SST}$. }
	\label{fig:Active}
\end{figure}

The $\boldsymbol{BCTS}$ is a set of configurations which contains exactly one unique configuration for each admissible torque vector $\ve{\tau}$. Fig~\ref{ActiveJ} illustrates the $\boldsymbol{BCTS}$~(green area) for the 2R planar robot~(cf.~Fig.~\ref{2D}). 
As configurations outside the $\boldsymbol{BCTS}$ are symmetric to those inside the $\boldsymbol{BCTS}$, ISMs can be learned for the \emph{entire} configuration space by exploring merely the $\boldsymbol{BCTS}$ and generalizing to the whole configuration space by exploiting the relations between symmetries. This restriction to the $\boldsymbol{BCTS}$ decreases the number of required samples by a factor of $8$ for the 2R planar robot when exploiting primary symmetries only. 

The $\boldsymbol{SST}$ was estimated for the 2R planar robot without any joint constraints as illustrated in Fig.~\ref{ActiveSOST}. The green area represents the torques corresponding to the $\boldsymbol{BCTS}$ in the $\boldsymbol{SST}$.

\subsubsection{Exploiting Symmetries with Modified Direction Sampling for the 2R Planar Robot - Online Learning}
\begin{figure}[!tb]
	\centering
	\includegraphics[height=50mm,width=0.5\textwidth]{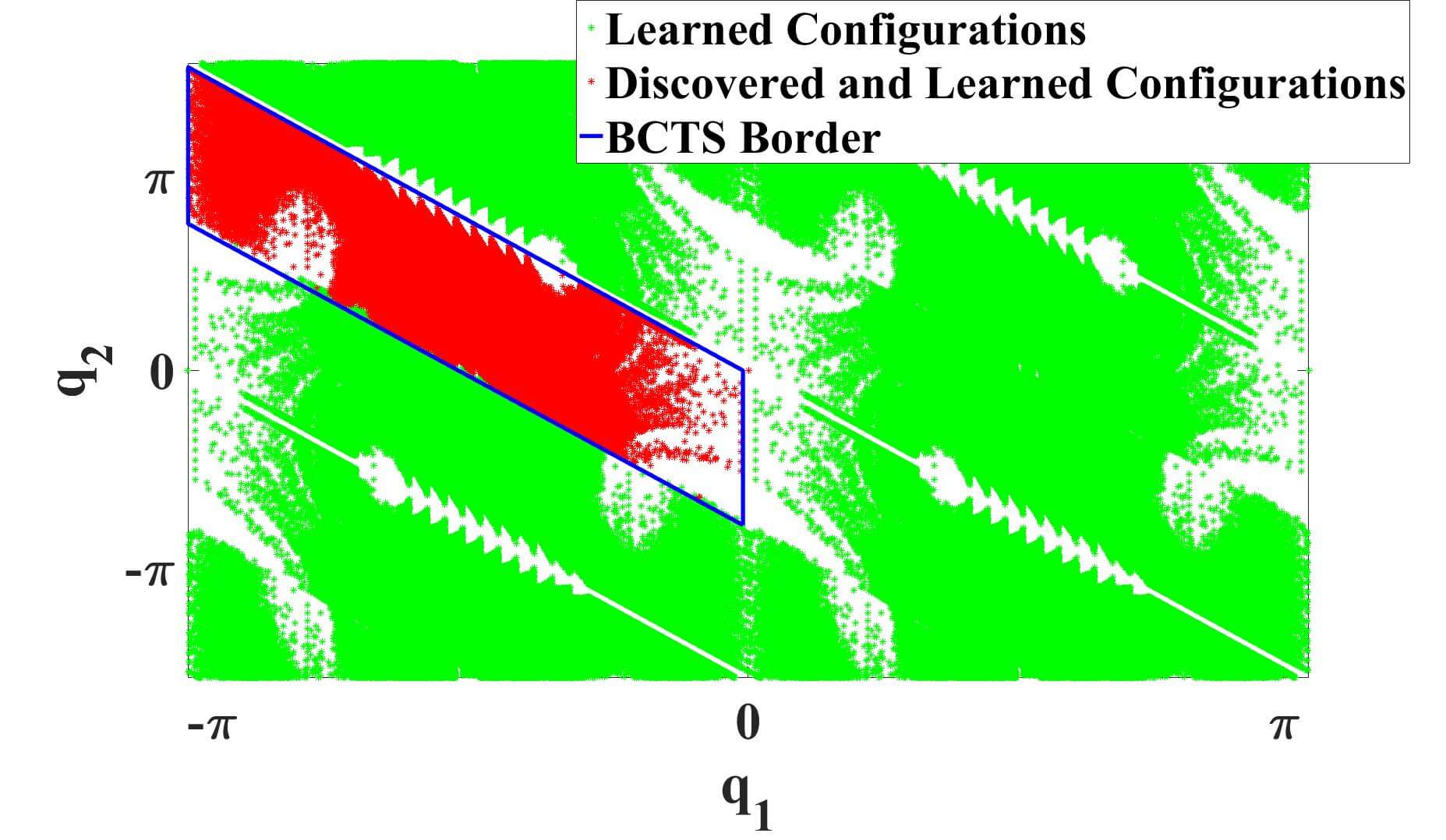}
	\caption{Explored configurations~(red) and learned configurations~(green).}
	\label{fig:Ex_Area}
\end{figure}

\begin{figure}[!tb]
	\centering
	\subfigure[Result for targets in the $\boldsymbol{BCTS}$.]{\label{fig:2R1}
    \includegraphics[height=50mm,width=0.4\textwidth]{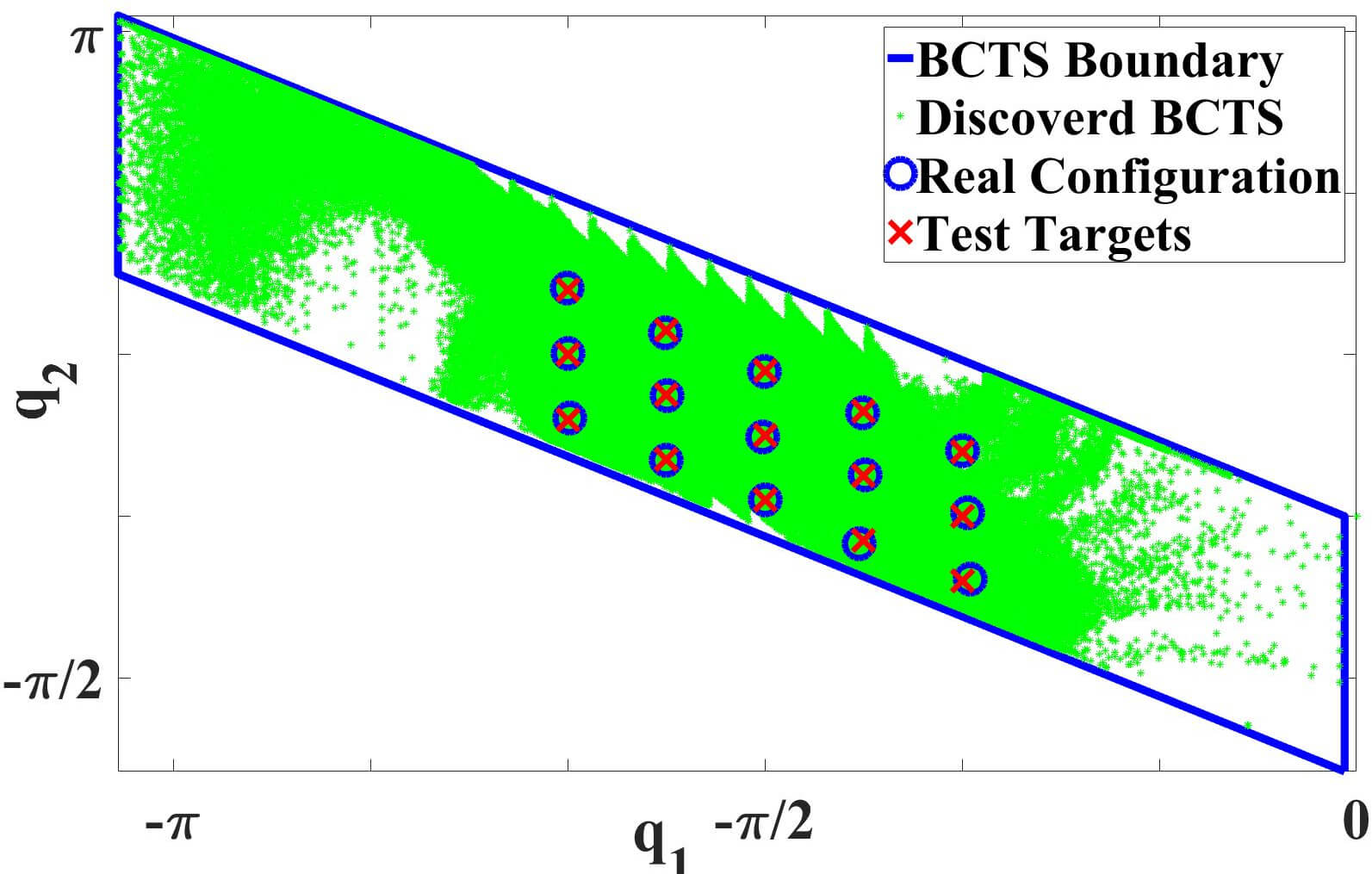}}
	\subfigure[Results for targets outside the $\boldsymbol{BCTS}$.]{\label{fig:2R2}
	\includegraphics[height=50mm,width=0.4\textwidth]{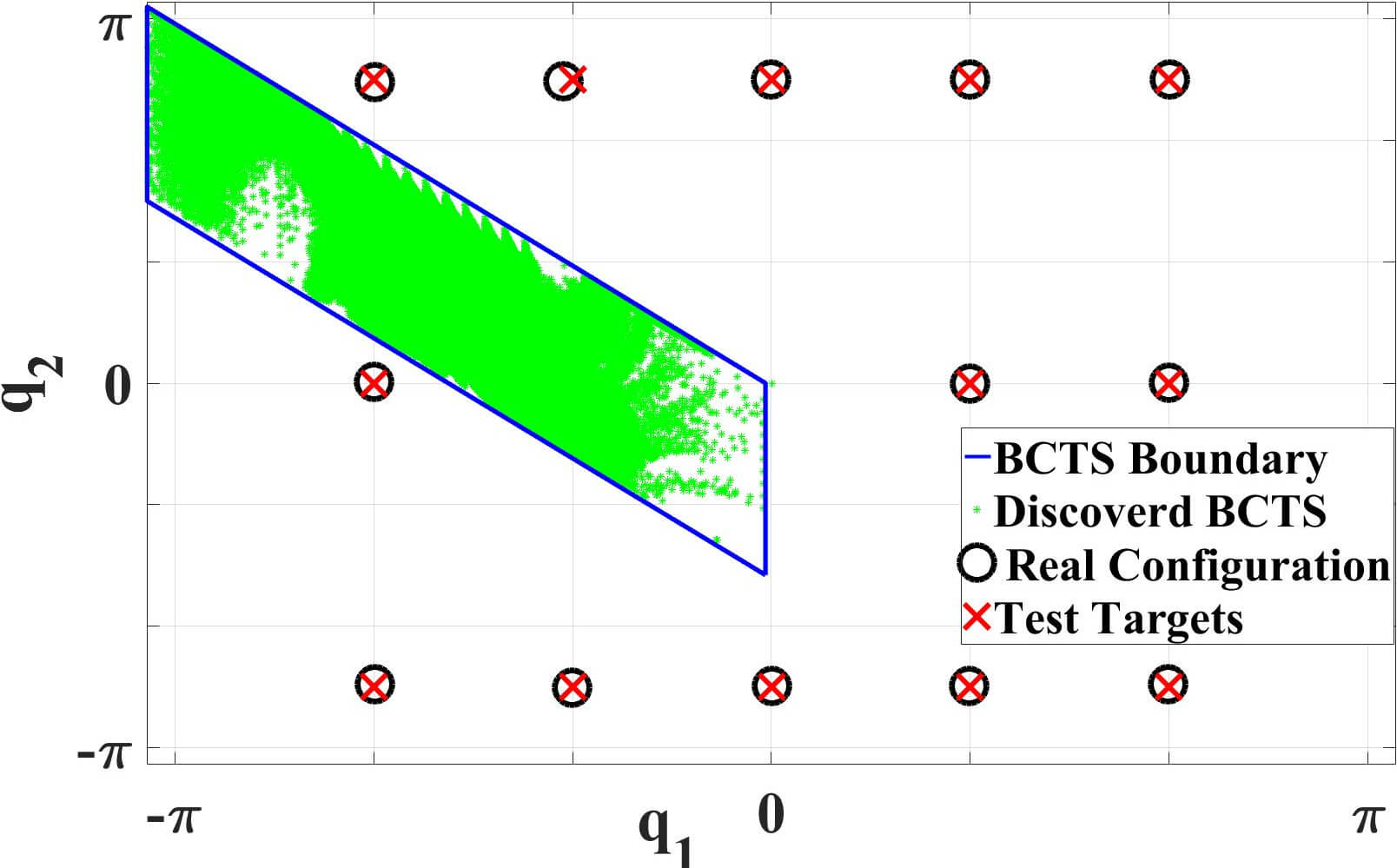}}
	\caption{Direction Sampling results for the 2R planar robot.}
	\label{fig:DS_result}
\end{figure}

Modified Direction Sampling was employed to discover the $\boldsymbol{BCTS}$ for the  2R planar robot~(cf.~Fig.~\ref{2D}). $3 \times 10^5$ samples were required to discover $82.197\%$ of the $\boldsymbol{BCTS}$. Fig.~\ref{fig:Ex_Area} shows the learned area of the configuration space~(green area) by exploring only the $\boldsymbol{BCTS}$ (red area). After the training phase, the robot was asked to reach and maintain $15$ configuration targets regularly distributed in the $\boldsymbol{BCTS}$. All targets were reached well with an RMSE of $4.5 mm$. The results are shown in Fig~\ref{fig:2R1}. The green area represents the discovered $\boldsymbol{BCTS}$, the boundary of $\boldsymbol{BCTS}$ is indicated by the blue parallelogram, red dots indicate the targets, and blue circles are the observed configurations.
Subsequently, the robot was asked to attain another $13$ targets scattered in the undiscovered area of the configuration space. The performance was also very good and the robot managed to achieve all targets with an RMSE of $2.8 mm$. As the test targets outside the $\boldsymbol{BCTS}$ are arranged on a regular grid as shown in Fig.\ref{fig:DS_result} and the targets inside the $\boldsymbol{BCTS}$ have a different arrangement, lower testing errors are plausible.
These results indicate that by exploiting symmetries in the ISM, it can be learned successfully by exploring only a very limited set of configurations.

\subsubsection{Exploiting Symmetries with Lattice Sampling for 3R Robot - Batch Learning}
\begin{figure}[!tb]
	\centering
	\includegraphics[height=50mm,width=0.5\textwidth]{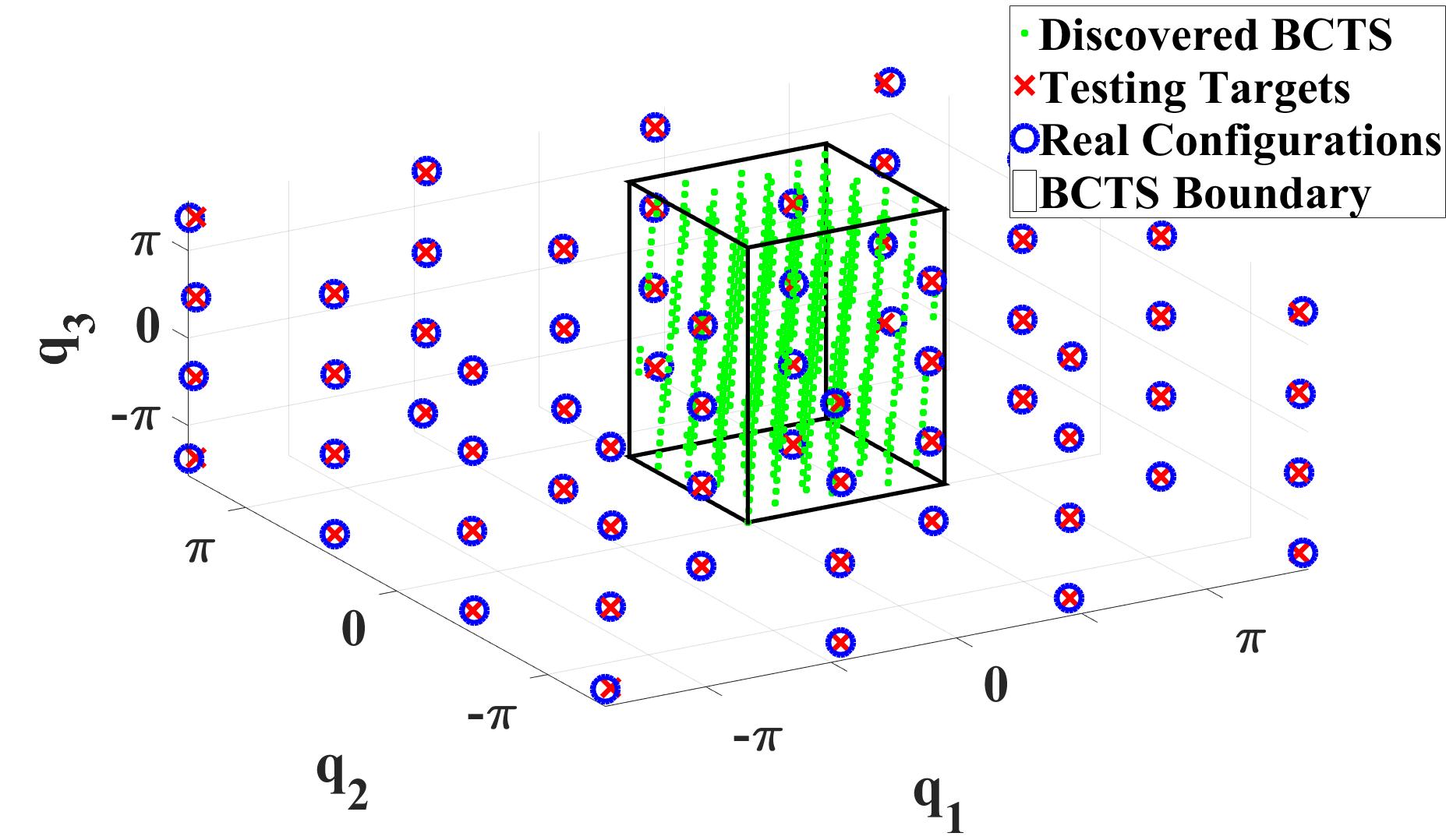}
	\caption{Batch learning result for the 3R manipulator based on lattice sampling and exploiting symmetries.}
	\label{fig:3R}
\end{figure}

To demonstrate the general applicability of the symmetry concept, batch learning to learn the ISM of the 3R robot~(cf.~Fig.~\ref{3D}) based on a lattice sampling approach was investigated. Lattice sampling~\cite{Sloan:85} was performed to collect training samples in the $\boldsymbol{BCTS}$. Gaussian noise with a standard deviation of $0.2Nm$ was added to the samples. A feed-forward neural network with one hidden layer consisting of $20$ neurons was used in a batch learning fashion. Only $700$ training samples in the $\boldsymbol{BCTS}$ were required to learn the ISM for the entire configuration space with a training RMSE of $0.04 Nm$. Then, the robot was asked to achieve $64$ targets arranged in regular grids and distributed over the entire configuration space. All targets were achieved very well as shown in Fig~\ref{fig:3R} with an RMSE of $0.0034 Nm$. The green area represents the discovered area in the $\boldsymbol{BCTS}$, the border of the $\boldsymbol{BCTS}$ is indicated by the cube, the testing targets are visualized by red crosses and the blue circles indicate the real configurations.
Lattice sampling was also performed for the entire configuration space. $11250$ samples were required to achieve approx. the same training RMSE~($0.041 Nm$) and the testing RMSE was $0.005 Nm$.

The number of required samples for $3R$ was reduced by a factor of $16$ by exploiting primary symmetries.
Hence, exploiting symmetries can drastically increase learning efficiency -- regardless whether offline or online learning schemes are considered -- by reducing the number of required samples by a factor which approx. equals the number of discovered primary symmetries. Further efficiency gains can be expected if secondary symmetries are exploited as well.




Note that the number of samples in batch learning is significantly lower than that required in the presented online learning approaches. Nevertheless, even batch learning approaches can greatly benefit from a significant reduction in the number of required samples by exploiting symmetries.
However, online learning techniques such as Goal Babbling and Direction Sampling, which generate targets on the fly and update the learner at each step simultaneously, best fit the concepts of gradual exploration as well as "learning while behaving" -- hence they best reflect human developmental aspects in robot learning.

\section{Conclusion and Outlook}
We showed that inverse statics mappings of various discretely-actuated serial manipulators can be learned accurately \emph{online} and \emph{from scratch} by modifying and extending the recently proposed online Goal Babbling and Direction Sampling schemes.

Moreover, we demonstrated that the efficiency of learning inverse statics mappings can be increased significantly by exploiting inherent symmetries of the mapping.
After discovering various sets of symmetric configurations for different motor torques, functional relations between the configurations were established and exploited in learning. 
To demonstrate the general applicability of our symmetry discovery and exploitation scheme, we successfully integrated it into our modified online Direction Sampling and a batch learning approach based on lattice sampling. The presented results indicated that factors of at least $8$ and $16$ w.r.t. the number of samples can be achieved for a 2R and a 3R robot resp.

Currently, the proposed Direction Sampling approach is being (i) enhanced to reduce the number of required samples, (ii) generalized to learn primary \emph{and} secondary symmetries for discretely-actuated serial manipulators with \emph{arbitrary} geometrical and inertial properties and (iii) extended to incorporate link and joint flexibility as well as nonlinear friction effects occurring in geared robots, which will pave the way for thorough experimental evaluation on a robot with variable stiffness actuators.

For motion planning and control, usually both the inverse kinematics and the inverse statics mapping are desirable.
Further developments will focus on combining inverse kinematics and inverse statics learning into one learning scheme to obtain both mappings in the same training phase.

\section*{ACKNOWLEDGMENT}
R. Rayyes receives funding from DAAD -\textquotedblleft Research Grants-Doctoral Programme in Germany" scholarship.



\bibliography{IEEEabrv,BiB2}

\bibliographystyle{IEEEtran}

\end{document}